%% file: bare_conf_compsoc.tex
\documentclass[conference,compsoc]{IEEEtran}
\usepackage[available,functional]{ieeebadges}

\input{header}

\input{macro}

\ifCLASSOPTIONcompsoc
  \usepackage[nocompress]{cite}
\else
  \usepackage{cite}
\fi
\ifCLASSINFOpdf
\else
\fi
\begin{document}
%
\title{\rev{\sys: Target Switch Attacks on Gimbal-Stabilized \\Visual Tracking Systems via Acoustic Injection}}

\author{
\IEEEauthorblockN{Jiarui Li}
\IEEEauthorblockA{University of Michigan\\
jiaruili@umich.edu}
\and
\IEEEauthorblockN{Joseph Brewington}
\IEEEauthorblockA{University of Michigan\\
brewing@umich.edu}
\and
\IEEEauthorblockN{Qingzhao Zhang\IEEEauthorrefmark{1}}
\IEEEauthorblockA{The University of Arizona\\
qzzhang@arizona.edu}
\and
\IEEEauthorblockN{Z. Morley Mao\IEEEauthorrefmark{1}}
\IEEEauthorblockA{University of Michigan\\
zmao@umich.edu}
}
\maketitle
\begingroup
\renewcommand\thefootnote{*}
\footnotetext[1]{These authors are corresponding authors.}
\endgroup

\input{sections/0_abst}

\begingroup
\renewcommand\thefootnote{\textcopyright~2026 IEEE}
\footnotetext{%
Personal use of this material is permitted.
Permission from IEEE must be obtained for all other uses.
This is the accepted version of the paper published in
2026 IEEE Symposium on Security and Privacy.
The final published version is available at IEEE Xplore.
DOI: 10.1109/SP63933.2026.00129.
}
\endgroup


%
\IEEEpeerreviewmaketitle

\input{sections/1_intr}
\input{sections/2_back}
\input{sections/3_thre}

\input{sections/4_desi}
\input{sections/5_eval}
\input{sections/6_disc}
\input{sections/7_conc}

\input{sections/8_ethi}





\bibliographystyle{IEEEtran}
\bibliography{ref}
\input{sections/9_appe}
%



\newpage 

\section{Meta-Review}

The following meta-review was prepared by the program committee for the 2026
IEEE Symposium on Security and Privacy (S\&P) as part of the review process as
detailed in the call for papers.

\subsection{Summary}
This paper presents \sys, an acoustic injection attack against UAV gimbal-stabilized visual tracking systems. The attack exploits MEMS gyroscope resonance to induce abnormal gimbal motion and camera viewpoint drift, which can cause visual trackers to lose the true target or switch to a nearby object. The paper develops an attack pipeline combining offline gimbal profiling with an online adaptive signal injection strategy. The evaluation includes both simulation and real-world experiments on a commercial drone platform, demonstrating how sensor-level disturbances can propagate through stabilization and affect application-level tracking behavior.

\subsection{Scientific Contributions}
\begin{enumerate}
\setcounter{enumi}{2}
\item Creates a New Tool to Enable Future Science.   
\item Provides a Valuable Step Forward in an Established Field.   
\item Establishes a New Research Direction.   
\end{enumerate}

\subsection{Reasons for Acceptance}
\begin{enumerate}
\item The paper demonstrates a practically relevant cross-domain attack path in autonomous systems, showing that acoustic injection at the sensing layer can propagate through gimbal stabilization and induce failures in visual tracking pipelines used by UAVs.
\item The work presents a complete attack workflow that combines offline profiling with an online adaptive attack loop, and evaluates the approach across multiple tracking algorithms and experimental settings.
\item The paper provides a useful experimental framework and methodology for studying perception-layer vulnerabilities in UAV systems, which can support follow-on research in robustness, detection, and mitigation.
\end{enumerate}

\subsection{Noteworthy Concerns} 
\begin{enumerate} 
\item \textbf{Interpretation of axis-dependent behavior}. The paper identifies axis-dependent responses (e.g., yaw-dominant behavior) through black-box profiling. While the empirical procedure is clearly described, it remains unclear whether such axis-selective resonance can be consistently identified across different gyroscope designs and implementations. For example, it is unclear whether similar yaw-dominant behavior can always be found for arbitrary commercial gimbal systems, or whether the observed behavior depends on specific hardware characteristics.
\item \textbf{Limited characterization of intermediate mechanisms}. The paper demonstrates that acoustic injection induces gimbal motion and leads to tracking degradation. However, the relationship between the induced oscillatory motion and the resulting tracking behavior is not fully characterized. In particular, it remains unclear how different motion patterns (e.g., oscillation versus accumulated drift) contribute to the observed tracking outcomes.
\item \textbf{Limitations of the in-flight evaluation}. The in-flight experiments demonstrate that acoustic injection can affect the built-in tracking system of a commercial drone. However, the evaluation is conducted under specific conditions (e.g., limited number of targets, controlled scenarios, and close-range acoustic injection), and does not explore how the attack behaves under more diverse real-world settings. As a result, the results should be interpreted within the scope of the evaluated scenarios.
\end{enumerate}



\end{document}

%% file: header.tex
\usepackage{algorithm}
\usepackage{algpseudocode}
\usepackage{soul}
\usepackage{graphicx}
\usepackage{filecontents}
\usepackage{diagbox}

\usepackage{tikz}
\usepackage{amsmath}

\usepackage{filecontents}

\usepackage{booktabs}    
\usepackage{multirow}    
\usepackage{subfigure}
\usepackage{subcaption}

\usepackage{xurl} 
\usepackage{amssymb}

\usepackage{colortbl}
\usepackage{array}
\usepackage{threeparttable}
\usepackage{gensymb}
\usepackage{xcolor}
\usepackage{xspace}

\usepackage{tcolorbox}
\tcbuselibrary{breakable,skins}  

\usepackage{hyperref}
\usepackage{makecell}

\newcommand{\rev}[1]{{\color{black} #1}} 
\newcommand{\revv}[1]{{\color{black} #1}} 

\usepackage{enumitem}
\setlist[itemize]{nosep, itemsep=1pt, topsep=1pt, leftmargin=1em, labelwidth=*}
\setlist[enumerate]{nosep, itemsep=1pt, topsep=1pt, leftmargin=1em, labelwidth=*}

\newcommand{\fullcircle}{%
  \begin{tikzpicture}[baseline=-0.5ex]
    \fill[black] (0,0) circle (0.35em);
    \draw (0,0) circle (0.35em);
  \end{tikzpicture}%
}

\newcommand{\emptycircle}{%
  \begin{tikzpicture}[baseline=-0.5ex]
    \fill[white] (0,0) circle (0.35em);
    \draw (0,0) circle (0.35em);
  \end{tikzpicture}%
}

\newcommand{\halfcircle}{%
  \begin{tikzpicture}[baseline=-0.5ex]
    \fill[black] (0,0) circle (0.35em);
    \fill[white] (0,0) -- (0.35em,0) arc (0:180:0.35em) -- cycle;
    \draw (0,0) circle (0.35em);
  \end{tikzpicture}%
}

\newtcolorbox[auto counter]{finding}[1][]{
  colback=black!5!white, colframe=black!75!black, boxrule=0.8pt, left=2pt, right=2pt, top=2pt, bottom=2pt,
  enhanced, drop shadow,
  breakable,
  fonttitle=\scshape,
  before upper=\textbf{Finding \thetcbcounter:\ },
  #1
}



%% file: macro.tex
\newcommand{\sys}{\texttt{Banshee}\xspace}

\newcommand{\droneModelOne}{HighEndDrone\xspace}
\newcommand{\droneModelTwo}{MidEndDrone\xspace}
\newcommand{\gimbalModel}{HighEndDrone's gimbal\xspace}
\newcommand{\trackAlgorithm}{HighEndDrone's tracking\xspace}
\newcommand{\sourceObj}{the true target\xspace}
\newcommand{\destinationObj}{the false target\xspace}

\newcommand{\myparagraph}[1]{\smallskip\noindent\textbf{#1}.\xspace}

\newcommand{\SwitchAtk}{Target switch\xspace}
\newcommand{\LossAtk}{Target loss\xspace}
\newcommand{\switchAtk}{target switch\xspace}
\newcommand{\lossAtk}{target loss\xspace}

%% file: sections/0_abst.tex
\begin{abstract}
Gimbal-stabilized visual tracking is critical for modern autonomous systems such as Unmanned Aerial Vehicles (UAVs).
While prior work shows acoustic signals can disturb gimbal internals, the impact of such attacks on real-world applications like UAV tracking and following remains underexplored.
Existing demonstrations largely overlook practical challenges for real-world attacks, such as object-motion uncertainty and runtime latency.
To bridge this gap, we present \sys\footnote{Our attack is named after the banshee, a mythical spirit whose scream causes or signals harm, reflecting the attack’s acoustic nature.}, the first physically realizable attack that \rev{induces target switching in UAV visual tracking systems} by exploiting acoustic vulnerabilities in gimbal-camera systems.
\rev{\sys generates carefully crafted acoustic waveforms that induce optimized adversarial gimbal oscillations, causing directionally biased camera-view drifts that break inter-frame target associations.}
\revv{Consequently, the onboard tracker is driven to switch from the original target to an attacker-selected object with high probability, with occasional target loss}.
\sys achieves a 93.6\% success rate in simulation across two commercial gimbal systems and five trackers. Real-world benchtop and in-flight black-box attacks against a commercial drone across varied scenarios show an overall 95.5\% attack success rate.
Our results reveal a practical cross-domain vulnerability between acoustics and vision, highlighting the need for robust designs of gimbal systems and applications. Our code is available at: https://github.com/U1ltra/Banshee. 
\end{abstract}

%% file: sections/1_intr.tex
\section{Introduction}
Gimbal-stabilized visual tracking is a core capability of modern camera systems, enabling persistent, high-precision object following in dynamic scenes. As a prominent example, commercial Unmanned Aerial Vehicles (UAVs) widely deploy target following, which typically pairs a multi-axis gimbal with an onboard camera, combined with object tracking algorithms running in software, to enable active tracking and following on a selected mobile target~\cite{han2021fast, cheng2017autonomous, maalouf2024follow, dji_followme_drone, skydio_followme_drone, autel_followme_drone}. Gimbal-stabilized visual tracking enables applications such as autonomous filming, surveillance, and infrastructure inspection, but also creates a single point of failure: compromising this pipeline can lead to severe consequences, including flight hazards, loss of vehicle control, and tracking of false targets~\cite{salameh2023federated, liwip, hollywood2025report, deception2025operation, davidson2016controlling}.

Gimbal systems commonly rely on real-time inertial measurement unit (IMU) feedback to mechanically stabilize onboard cameras during rapid motion~\cite{bereska2013system, altan2020model}. Prior research has shown that carefully crafted acoustic patterns, delivered via speakers, ultrasonic transducers, or even laser systems, can manipulate IMU readings and, in turn, disrupt gimbal stabilization~\cite{trippel2017walnut, tu2018injected, son2015rocking, gao2022kite, shamsiwip}. While these studies establish the feasibility of influencing gimbal motion at the sensor level, they remain disconnected from real-world applications such as UAV target tracking and following, leaving their practical impact uncertain.


\begin{figure}
    \centering
    \includegraphics[width=0.9\linewidth]{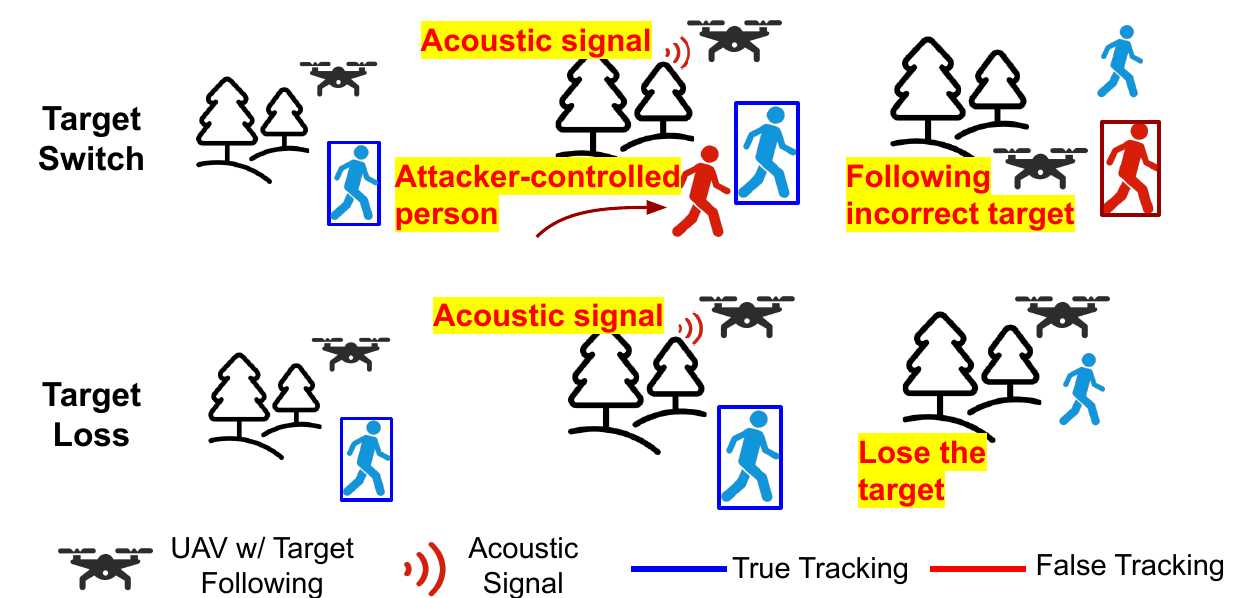}
    \caption{
    Illustration of \sys in UAV target following. Crafted acoustic signals induce the UAV's visual tracker to switch to an incorrect target or lose track.}
    \label{fig:attack_overview}
\end{figure}

To address this gap, we propose \sys, the first physical target switch attack against UAV visual tracking systems via acoustic injection, linking a hardware acoustic vulnerability with application-level tracking weaknesses to enable end-to-end system compromise.
\rev{By injecting acoustic signals to the UAV, the attack induces gimbal oscillations with directional bias that accumulates drift along a vulnerable axis. This abnormal motion disrupts motion smoothness and corrupts the tracker’s association with the true target, increasing the likelihood of a incorrect tracking output.}
Figure~\ref{fig:attack_overview} illustrates two scenarios. (1) In the \emph{\switchAtk} attack, a UAV performing target following switches from its original target to an attacker-selected object, enabling the attacker to potentially steal the UAV from its owner or transfer tracking onto an unintended target. (2) In the \emph{\lossAtk} attack, the UAV loses its tracking target, allowing a suspect under surveillance to escape.

The attack has two stages. In \emph{offline gimbal profiling}, \rev{the attacker uses an identical gimbal to learn a black-box mapping from acoustic signals to induced gimbal motion}. In the \emph{online attack}, the attacker runs two loops simultaneously. 
A \emph{surrogate tracking} loop that runs a surrogate of the UAV onboard tracking mimicking the actual UAV tracking behavior using black-box knowledge. A \emph{planning-execution} loop then optimizes a sequence of acoustic signals under physical and algorithmic constraints, leveraging both the gimbal acoustic response model and the tracking surrogate. 
\rev{These signals are then injected through a speaker or piezoelectric transducers to induce the desired gimbal motion, probabilistically biasing the tracker away from the true target toward incorrect associations.}

Designing this application-aware acoustic attack raises several key challenges. 
\rev{First, the attack must achieve an empirically sufficient alignment between physical acoustic signals and their induced camera motion, in order to produce desired adversarial motion that disrupts tracking.}
Second, the attack must operate at runtime without prior knowledge of the UAV behavior, which motivates an adaptive online strategy that updates the surrogate tracker and signal injection plan as the scene evolves. 
Third, the attack must remain effective under real-world uncertainties, including unknown future object motion, which we address with optimization algorithms that tolerate uncertainties. 
\revv{\sys overcomes these challenges and achieves high-probability empirical success in inducing \switchAtk or \lossAtk.}

Extensive experiments prove the practicality of the proposed attack. First, offline profiling on built-in gimbals of two commercial UAV models shows that \rev{consistent runtime directional bias in gimbal motion} is feasible. Second, we run large-scale simulation in Gazebo simulator, which deploys PX4-Autopilot flight stack, uses the profiled gimbal parameters, and tests the attack on five representative trackers and diverse scenarios. The results show that \sys overall corrupts the tracking in 93.6\% of trials (including 75.0\% \switchAtk and 18.6\% \lossAtk), proving attack effectiveness and robustness. Finally, real-world experiments further validate successful black-box \switchAtk on commercial drones, including a realistic exploit on the built-in object tracking during flight.\footnote{We have completed responsible disclosure to the vendor and, at their request, anonymized the commercial product models. This anonymization does not affect reproducibility; the vulnerability is not specific to the vendor’s product and may apply broadly to gimbal mechanisms, and we provide proof-of-concept simulation environments at \href{https://github.com/U1ltra/Banshee}{GitHub}.}
We summarize our contributions:
\begin{itemize}
\item We design \sys, \rev{the first end-to-end attack that uses acoustic injection to compromise UAV visual tracking, bridging gimbal-level vulnerabilities to application-level impacts, and adapts to diverse real-world conditions.}

\item We propose \rev{the first systematic method to empirically model gimbal motion under acoustic injection}. Using offline profiling and online phase modulation, the attacker can \rev{induce consistent, directionally biased gimbal angular offsets in real time.}

\item We extensively evaluate \sys in high-fidelity simulation (Gazebo + PX4-Autopilot), with realistic simulation of gimbal vulnerability. We also demonstrate real-world online attacks on a commercial \droneModelOne.

\end{itemize}

%% file: sections/2_back.tex
\section{Related Works}

\begin{table}[t]
    \centering
    \renewcommand{\arraystretch}{0.1} 
    \setlength{\tabcolsep}{1.0pt} 
    \scriptsize
    \begin{threeparttable}
        \begin{tabular}{|c|c|c|c|c|c|c|}
            \cline{1-5}
            Acoustic Attack & \makecell{\textbf{\rev{Consistent}}\\\textbf{\rev{Biasing}}} & \makecell{\textbf{Affected}\\\textbf{Module}} & \textbf{Signal Source} & \makecell{\textbf{Evaluation}\\\textbf{Platform}}  \\
            \cline{1-5}
            \textbf{Rocking.}\cite{son2015rocking}  & \emptycircle & Firmware & Speaker & Drone controller  \\ 
            \textbf{WALNUT}\cite{trippel2017walnut}& \fullcircle  & Sensor & Speaker & Toy car  \\ 
            \textbf{Injected}\cite{tu2018injected}  & \fullcircle  & Firmware & Speaker & Smart phone  \\ 
            \textbf{Poltergeist}\cite{ji2021poltergeist}  & \emptycircle  & Detection & Speaker & Smart phone  \\ 
            \textbf{KITE}\cite{gao2022kite}  & \fullcircle  & Localization & Speaker\&Piezo & Drone controller \\ 
            \textbf{Laser}\cite{shamsiwip}  & \emptycircle  & Detection & Laser & Drone gimbal   \\ 
            \textbf{Ours}  & \fullcircle  & Tracking & Speaker\&Piezo & Drone gimbal  \\ 
            \cline{1-5}
            Tracking Attack & \makecell{\textbf{Online}\\\textbf{Optimize}} & \makecell{\textbf{Real-world}\\\textbf{Robustness}} & \makecell{\textbf{Attack}\\\textbf{Vector}} & \textbf{Impact/Algo.} \\
            \cline{1-5}
             \textbf{Fooling}\cite{jia2020fooling} & \emptycircle & \emptycircle & Patch & Loss/MOT  \\
             \textbf{AttrackZone}\cite{muller2022physical} & \fullcircle & \emptycircle & Projector & Loss/SOT  \\
             \textbf{ControlLoc}\cite{ma2024controlloc} & \emptycircle & \halfcircle\tnote{\ddag} & Patch & Loss/MOT  \\
             \textbf{AdvTraj}\cite{wangphysical} & \fullcircle & \emptycircle & Person & Switch/MOT  \\
             \textbf{FlyTrap}\cite{xie2025flytrapphysicaldistancepullingattack} & \emptycircle & \halfcircle\tnote{\ddag} & Patch & Shrink/SOT  \\
             \textbf{Ours} & \fullcircle & \fullcircle & Acoustic & Switch,Loss/SOT,MOT \\
            \cline{1-5}
        \end{tabular}
        \begin{tablenotes}
            \item[\ddag] \scriptsize Robust patch generation by applying image transformations.  
        \end{tablenotes}
    \end{threeparttable}
    \caption{Comparison with prior acoustic/tracking attacks.}
    \label{tab:related}
\end{table}

\myparagraph{UAV visual tracking systems}
Visual tracking enables core UAV functions such as target following, obstacle avoidance, and aerial monitoring~\cite{dji_followme_drone, ji2022elastic, maalouf2024follow, gao2023adaptive, liu2022multi}. Tracking algorithms generally fall into single-object tracking (SOT) and multi-object tracking (MOT).
SOT methods, widely used for target following, include correlation-filter-based~\cite{li2020autotrack, ye2021multi} and siamese-network-based approaches~\cite{xing2022siamese, cao2022tctrack}, both relying on appearance similarity. More advanced methods (e.g., transformer-based~\cite{yan2021stark} and online learning-based~\cite{danelljan2018atom, bhat2019dimp}) are often too resource-intensive for onboard deployment.
MOT methods, used in aerial surveillance and obstacle avoidance~\cite{liu2022multi, du2021giaotracker, dji_followme_drone}, maintain associations across multiple dynamic objects and largely rely on motion consistency.
Across these systems, a shared assumption is smooth inter-frame object displacement under benign conditions—an assumption our attack exploits.

\myparagraph{Acoustic vulnerabilities of gimbal systems}
Gimbal-stabilized cameras are standard in UAVs for stable video capture~\cite{dji_followme_drone, skydio_followme_drone, autel_followme_drone}, relying on IMU-based feedback control~\cite{bereska2013system, altan2020model}. Prior work (Table~\ref{tab:related}) has shown that MEMS IMUs are vulnerable to acoustic injection~\cite{trippel2017walnut, tu2018injected, son2015rocking, gao2022kite, shamsiwip, jeong2023rocking}. By exploiting resonant frequencies, attackers can inject false IMU signals, causing the gimbal to compensate for nonexistent motion and shift camera orientation.
While these attacks demonstrate destabilization or sensor spoofing, they have not been systematically leveraged to induce \switchAtk on visual tracking systems.

\myparagraph{Prior physical attacks on visual tracking}
Physical attacks on visual tracking have been explored but face unique challenges in UAV settings (Table~\ref{tab:related}). 
Adversarial patch–based attacks~\cite{jia2020fooling, ma2023wip, ma2024controlloc} can degrade tracking, yet lack online adaptation~\cite{liu2023rpau, wiyatno2019physical}
Real-time attacks using projectors or adversarial trajectories~\cite{wangphysical, muller2022physical} struggle with real-world uncertainty and latency. Acoustic attacks on object detection~\cite{ji2021poltergeist, 285403} do not extend to the more complex tracking pipeline. FlyTrap~\cite{xie2025flytrapphysicaldistancepullingattack} disrupts tracking via physical patches but under a different threat model.
To date, no prior work demonstrates a physically realizable acoustic \switchAtk attack on UAV visual tracking.

%% file: sections/3_thre.tex
\begin{figure}[t]
    \centering
    \includegraphics[width=\linewidth]{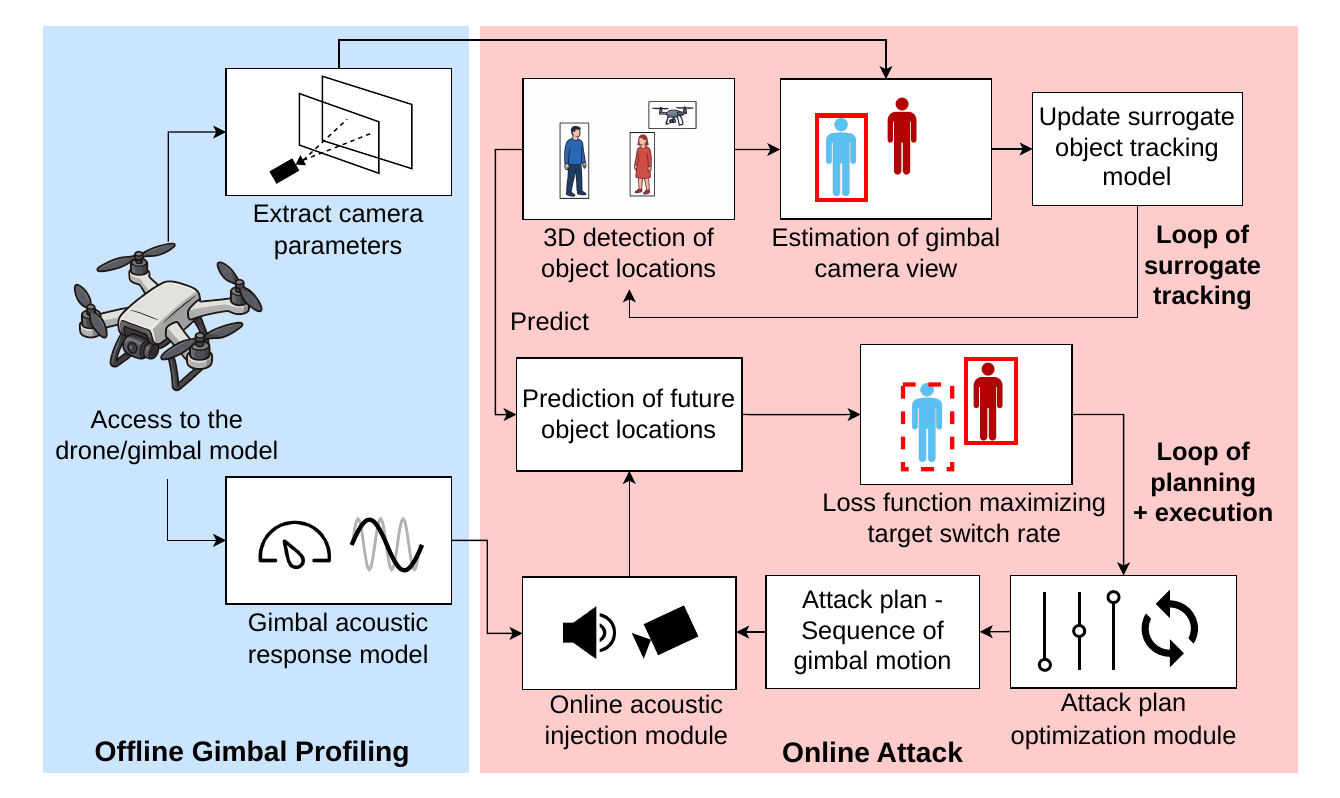}
    \caption{\rev{Overview of \sys attack.}}
    \label{fig:attackdiag}
\end{figure}

\section{Threat Model}
\label{sec:threat_model}

\myparagraph{System and environment assumptions}
In the attack scenario, a UAV performs target following using a gimbal-stabilized camera, whose video stream is processed by an onboard visual tracker. The flight controller then commands the UAV to follow the tracked target. \rev{Besides the legitimate target, the scene contains an attacker-controlled object that can be placed near the target during the attack.} 
For example, if the UAV is following a pedestrian, an attacker-guided pedestrian can casually walk nearby (e.g., within a few meters), creating a normal and inconspicuous scenario. \rev{We consider a setting with no other nearby objects, as commonly seen in open-space tracking, and leave attacks in crowded scenes to future work.}
We denote the originally tracked object as \textbf{\sourceObj} and the object after switching as \textbf{\destinationObj}. 

\myparagraph{Attack goal}
The adversary aims to launch an online \textbf{\switchAtk} attack that achieves persistent target switching from \sourceObj to \destinationObj. 
The attack succeeds if the tracker consistently associates \destinationObj, within the same semantic category (e.g., pedestrian or vehicle), as \sourceObj. 
In different application contexts, the attacker can achieve two representative outcomes:
(1) \emph{Takeover}: the UAV, following its legitimate target (e.g., its owner), switches to an attacker-controlled object, resulting in effective capture;
(2) \emph{Escape}: the UAV, surveilling an attacker-controlled target, switches to another object, allowing the attacker to evade tracking.
Both are instances of \switchAtk, and we therefore do not distinguish them in evaluation.

A secondary goal is the \textbf{\lossAtk}, where the tracker loses the target entirely, resulting in denial of service. While \lossAtk often arises when \switchAtk fails, it still has significant adversarial impact.

\rev{Due to real-world uncertainties and limited attacker knowledge (discussed below), the attack is inherently probabilistic rather than deterministic; nevertheless, it should achieve a high probability of tracking corruption.}

\myparagraph{Attacker knowledge and capability}
First, we assume the attacker knows the target gimbal-camera model and can obtain an identical device, which is feasible for commercially available UAVs. Using this device, the attacker performs offline gimbal profiling (Section~\ref{sec:gimbal-profile}) to characterize its acoustic vulnerability and learn an empirical mapping from acoustic signals to induced gimbal motion, which supports the online attack (Section~\ref{sec:online_attack}). \rev{This mapping guides not only signal optimization but also attacker object placement, e.g., aligning with the most vulnerable gimbal axis and maintaining a moderate distance from \sourceObj.} The attacker also obtains key system parameters, including gimbal constraints (e.g., rotation limits and degrees of freedom) and the camera intrinsic matrix.

Second, we assume the attacker has only black-box knowledge of the tracking algorithm deployed on the UAV system. The attacker does not know any implementation details including the algorithm and parameters, but the attacker can exploit surrogate models to mount transfer attacks.

Third, the attacker needs a mechanism to deliver acoustic impact during flight. The most reliable option is attaching a \textbf{malicious payload} (such as a piezoelectric transducer~\cite{gao2022kite}) to ensure stable acoustic signal injection. Installing a malicious payload requires physical access, for example, an attacker posing as a maintenance technician could covertly attach the device during routine UAV servicing. 

Finally, the attacker can observe the UAV system, \sourceObj, and \destinationObj throughout the attack, allowing a sensor system (such as LiDAR, radar, or camera) to perform \textbf{real-time 3D object detection} on the UAV system and key objects to be tracked, obtaining location information necessary for runtime attack optimization. The attacker also needs to know the \textbf{real-time gimbal orientation} to inject biased gimbal motion (Section~\ref{sec:gimbal-profile}) and achieve an accurate online surrogate of UAV tracking (Section~\ref{sec:loop_surrogate}). To this end, the attacker can install an IMU on the gimbal (as part of the malicious payload).
Note that we do not assume any access to the UAV system's camera stream or internal software data. Our attack predicts the behavior of the tracking algorithm relying only on the external sensing and the profiled gimbal system parameters. 

\myparagraph{Relaxed requirements}
\rev{Beyond the \textbf{primary} threat model, we consider a \textbf{relaxed} setting that reduces attacker requirements and validate the effectiveness of this threat model through physical experiments (Section~\ref{sec:eval-phy}). 
To avoid physical access, the attacker may perform a \textbf{remote attack} via long-range ultrasonic emitters~\cite{ji2021poltergeist, longRangeUltra} or laser-induced vibrations~\cite{shamsiwip} (see Section~\ref{sec:discussion}). The attacker may also estimate gimbal orientation using vision-based methods~\cite{yang2019scrdet, xu2023dynamic} instead of installing a malicious IMU. 
}


%% file: sections/4_desi.tex
\section{\sys Design}

Figure~\ref{fig:attackdiag} shows the two stages of \sys: offline gimbal profiling and online attack.

\textbf{Offline gimbal profiling} requires access to a gimbal identical to the target. \rev{Through controlled acoustic injection experiments, the attacker infers camera parameters, identifies the most vulnerable gimbal axis, and learns a black-box mapping from acoustic signals to induced motion (Gimbal Acoustic Response Model). This stage guides the selection of attack signals that induce abnormal gimbal motion.} Details are in Section~\ref{sec:gimbal-profile}.

\textbf{Online attack} injects crafted acoustic signals into the target gimbal to induce abnormal motion along a vulnerable axis and cause tracking errors such as \switchAtk. It consists of two iterative loops: \emph{surrogate tracking} and \emph{planning-execution}. The surrogate tracking loop maintains a surrogate tracker that mimics the UAV tracker, updated using the attacker’s external sensor data and the estimated gimbal view. The planning-execution loop optimizes a sequence of acoustic signals under physical and algorithmic constraints, using the offline profiling results, the surrogate tracker, and the current 3D scene as inputs. These signals are then transmitted through the attack vector (for example, speakers and piezoelectric transducers) to gradually redirect the tracking from the legitimate target to the attacker-selected object. Details are in Section~\ref{sec:online_attack}.

\subsection{Adversarial Acoustic Gimbal Biasing}
\label{sec:gimbal_control}

\label{sec:gimbal-profile}
\begin{figure}[t]
    \centering
    \includegraphics[width=\linewidth]{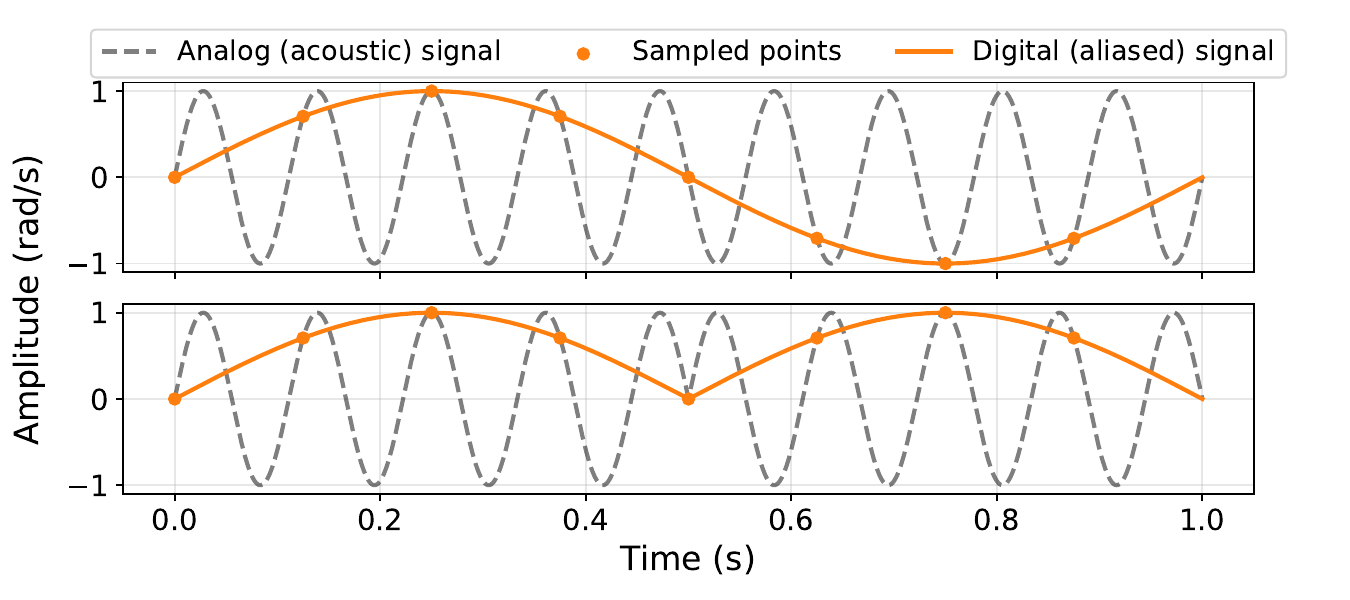}
    \caption{Illustration of phase-based directional bias in gimbal motion. Injected and aliased signals with (top) and without (bottom) phase shifting.}
    \label{fig:phase_shift}
\end{figure}


\rev{We design a physical attack that injects selected acoustic waveforms \emph{at runtime} to induce biased, abnormal gimbal motion. Specifically, the acoustic signal excites resonance and corrupts gyroscope readings, causing the stabilization controller to apply incorrect compensation. This results in \emph{directionally biased oscillations} that accumulate into abnormal gimbal motion.}

Formally, by offline measurements on the the gimbal device, we \rev{empirically approximate} a mapping from acoustic waveform with frequency \(f_{\mathrm{in}}\) and amplitude \(A_{\mathrm{in}}\) to the gimbal motion response as a three-axis angular velocity \(\boldsymbol{\omega}[t]\). Let the acoustic drive be \(x(t)=A_{\mathrm{in}}\sin(2\pi f_{\mathrm{in}}t)\). We learn a \textbf{Gimbal Acoustic Response Model} \(\hat{\mathcal{M}}\) satisfying
\begin{equation}
\boldsymbol{\omega}[t] \approx \hat{\mathcal{M}}\!\big(A_{\mathrm{in}},\,f_{\mathrm{in}},\,t\big) + \boldsymbol{\varepsilon}[t],
\end{equation}
where \(\boldsymbol{\varepsilon}[t]\) is the residual error.
This section presents the theoretical background and the offline profiling procedure used to obtain \(\hat{\mathcal{M}}\), detailed in Sections \ref{sec:digital-to-control}-\ref{sec:profiling-steps}.
In addition, runtime \rev{directional biasing} must handle uncertainty in the gyroscope sampling phase, which affects the induced motion direction; we therefore apply a phase-modulation routine that estimates and aligns the sampling phase at runtime, as detailed in Section \ref{sec:gimbal-control-online-steps}.


\subsubsection{Gimbal Fundamentals: Analog-Digital-Control} 
\label{sec:analog-to-digital}


MEMS gyroscopes measure angular velocity via the Coriolis force on a vibrating proof mass. Suspended by springs, the mass resonates at its natural frequency $f_n$. In previous work, it has been established that an external acoustic signal with frequency $f \approx f_n$ can drive large oscillations even without actual motion~\cite{son2015rocking, trippel2017walnut, tu2018injected}. We use the definitions standard among works on acoustic injection attacks and model the response of the proof mass as a driven harmonic oscillator:
\begin{equation}
\label{eq:analog}
    \omega(t) = A \sin(2\pi f t + \phi),
\end{equation}
where $A$ is the vibration amplitude and $\phi$ is the phase of the signal.

These oscillations are then sampled by an analog-to-digital converter (ADC). Since the sampling rate $f_s$ is much lower than the resonant frequency, the digitized signal appears as a lower-frequency alias rather than at $f$ (Figure \ref{fig:phase_shift}). For an ideal, constant-rate ADC, the sampled signal is
\begin{equation}
\label{eq:digital}
    \omega[t] = A \sin(2\pi f_d \frac{t}{f_s} + \phi),\; \{t \in \mathbb{N}\}.
\end{equation}
The digitized frequency of the aliased signal is subject to the Nyquist theorem
\begin{equation}
\label{eq:digifreq}
    f_d = f - n \cdot f_s \;\; \{n\in\mathbb{N},\;\; f_d \leq \frac{1}{2}f_s\}.
\end{equation}
The aliased frequency $f_d$ is always less than half of $f_s$ and approaches zero as the frequency of the injected signal nears an integer multiple of this rate. 

The digitized acoustic signal propagates into the system as perceived motion, leading to compensatory control algorithm behaviors \cite{STorM32BGC, BaseCam}.
Because commercial IMUs operate at modest sampling frequencies typically in the tens to hundreds of hertz and rarely exceeding 1~kHz~\cite{invenSense, stmicro, bosch,stmicro2}, the resulting aliased oscillations appear at low, easily observable frequencies, enabling our proposed profiling and modeling methodology described below.

\subsubsection{The Model of \rev{Adversarial Gimbal Biasing}}
\label{sec:digital-to-control}
In theory, an adversary can \rev{influence and bias gimbal motion through modulated acoustic signals. To approximate this goal, we design a simplified gimbal motion abstraction} leveraging structural and algorithmic insights as follows.

\myparagraph{Scaling vibration amplitude}
Amplitude scaling leverages the linear relationship between the injected signal amplitude and the resulting vibration amplitude. Following the amplitude equation for a driven harmonic oscillator we have
\begin{equation}
\label{eq:amplitude}
\revv{A = \frac{1}{m\;Z_m\;2 \pi f} \cdot A_{in},}
\end{equation}
where the specific IMU model determines mass $m$ and mechanical impedance $Z_m$, while the attacker controls signal frequency $f$ and the acoustic amplitude $A_{in}$. 
We capture this linear relationship by fitting a linear regression for each of the identified resonant frequencies, simplifying the denominator to a single constant $a$
\begin{equation}
\label{eq:amplitude2}
\revv{\omega[t] = (a \cdot A_{in}) \sin(2\pi f_d \frac{t}{f_s} + \phi).}
\end{equation}

\myparagraph{\rev{Correcting} direction of gimbal motion}
The direction of gimbal motion is subject to real-world uncertainty since the phase of $\omega[t]$ depends on the relative timing between the injected signal and the sampling of the gimbal gyroscope. We extend the phase modulation technique in~\cite{trippel2017walnut} to address limitations when the timing of samples is subjected to drift.
We modulate phase in response to the direction of detected gimbal motion to improve robustness in the presence of sampling rate drifts. Specifically, when the direction of the observed gimbal motion begins to differ from the intended motion, we shift the phase of the injected signal by $\pi$ so as to invert $\omega$, also shown in Figure \ref{fig:phase_shift}
\begin{equation}
\label{eq:phase-shift}
    -\omega[t] = A \sin(2\pi f_d \frac{t}{f_s} + \phi + \pi)
\end{equation}

\myparagraph{\rev{Adversarial biasing} is independent of the gimbal orientation at the time of signal injection} 
UAV gimbal systems exhibit significant variability in servo actuation as a response to injected acoustic signals depending on gimbal orientation. The cause of this phenomenon lies in the gimbal stabilization mechanism. A 3-axis gimbal stabilizes a mounted camera using servos fixed in the drone body reference frame, while the camera body houses a gyroscope that is used in a feedback control loop to maintain stability in the world reference frame. To achieve this, gyroscope readings must be rotated by the orientation of the camera body relative to the drone body in order to actuate the correct servos.
The closed-form matrix derived in~\cite{rajesh2015gimbal} describes the rotation from the drone reference frame to the camera body reference frame that is used to transform gyroscope readings into correct servo actuation. Given angular values of pitch, roll, and yaw $\mathbf{\theta}$, the rotation matrix is
\begin{equation}
\label{eq:rotation}
R = R_{p} R_{r} R_{y} =
\left[\begin{smallmatrix}
C_y C_p - S_y S_r S_p & - C_r S_y & C_y S_p + C_p S_y S_r \\
C_p S_y + C_y S_r S_p & C_y C_r & S_y S_p - C_y C_p S_r \\
- C_r S_p & S_r & C_r C_p
\end{smallmatrix}\right]
\end{equation}
where $C_i = \cos{\theta_i}$, $S_i = \sin{\theta_i}$, and current gimbal orientation $\theta$.
In short, when motion is perceived in the gimbal gyroscope the stabilization system will always actuate the correct servos in order to produce an opposite motion along the same axis in the camera-body frame, resulting in a stable motion response independent of the gimbal orientation. Section \ref{sec:eval-gimbal} also shows supporting results.


Overall, our Gimbal Acoustic Response Model \(\hat{\mathcal{M}}\) has the following form
\begin{equation}
\label{eq:adv-control}
\revv{\omega[t] = a \cdot A_{\text{in}}\,}
\begin{cases}
 \sin\!\big(\theta_t\big), & \text{if } \operatorname{sgn}\!\big(\sin\theta_t\big)=s,\\[4pt]
 \sin\!\big(\theta_t+\pi\big), & \text{if } \operatorname{sgn}\!\big(\sin\theta_t\big)\neq s,
\end{cases}
\end{equation}
where $\theta_t = 2\pi f_d \frac{t}{f_s} + \phi$ and intended direction $s\in\{-1,1\}$. In the following section, we present the profiling steps to obtain $f_d$ and $a$. The additional residual error term \(\boldsymbol{\varepsilon}[t]\) is modeled as Gaussian noise, parameterized using the real-world experimental traces.

\subsubsection{Gimbal Profiling for Obtaining the Model} 
\label{sec:profiling-steps}

The profiling procedure consists of two major steps that sweep over a single selected parameter of interest -- \textbf{frequency} and \textbf{amplitude}, which produces $f_d$ and $a$, respectively.
In each sweep we first collect raw profiling data, then proceed to extract key information using spectral analysis and construct the necessary model. 

\myparagraph{Black-box gyroscope modeling via observation}
Without loss of generality, we consider a three-axis gimbal camera system with an embedded three-axis gyroscope. By applying acoustic signals, the attacker induces oscillating motion readings in each axis, $\omega[t] = [\omega_p[t], \omega_r[t], \omega_y[t]]^T$, resulting in observable gimbal motion. 
The attacker installs an external malicious gyroscope on the camera body and collects the external gimbal motions $\omega_{obs}[t]$.

\myparagraph{Frequency sweep}
The first profiling step returns the set of resonant frequencies $Q$ of the gyroscope that are most effective in driving proof-mass oscillations.
(1) Following~\cite{tu2018injected}, sweep a sine wave across a wide range of acoustic signal frequencies (e.g., 1~Hz to 40~kHz) and record the motion response of the gimbal. 
(2) The acoustic amplitude $A_{in}$ should be set to a fixed maximum value. For each injected test frequency $f$, extract aliased frequency $f_d$ and amplitude $A$ using spectral analysis.
(3) Examining amplitude as a function of injected frequency and identifying local maxima of $A(f)$; these peaks reveal ideal injected signal frequencies for attacks and we collect them into a set $Q$.

\myparagraph{Amplitude sweep}
The second profiling step returns the linear relationships $\mathcal{A}$ between the injected acoustic amplitude and the amplitude of the induced gimbal motion.
(1) For a selected frequency in $Q$, sweep the injected acoustic signal amplitude $A_{in}$ from low to high (e.g., 1\% to 100\% signal power) and record the motion response of the gimbal.
(2) For each injected signal, extract the observed motion amplitude $A$ using spectral analysis.
(3) Fit a linear regression of $A$ as a function of $A_{in}$ to obtain a constant $a$ such that $A \approx a \cdot A_{in}$.
(4) Repeat the procedure for remaining frequencies in $Q$.



\subsubsection{\rev{Directional Biasing} During Online Attacks} 
\label{sec:gimbal-control-online-steps}
Since the phase of the sampled signal cannot be determined during offline profiling, we bias the direction of gimbal motion with real-time feedback using the theoretical model in Section \ref{sec:digital-to-control}. For enhanced robustness against real-world uncertainties, we can use a history-based method and only trigger a switch when the past $N$ samples show motion in the incorrect direction. We also incorporate a time restriction, limiting the number of phase switches to a period of once per elapsed time $T$. The attacker can tune $N$ and $T$ to allow the proper level of feedback sensitivity. As a general rule of thumb, $N < f_{\text{feedback}}$ and $T < \frac{1}{f_d}$, where $f_{\text{feedback}}$ is the frequency of the motion direction detection. 


In summary, the attacker can \rev{inject consistent and directionally biased gimbal angular offsets in real-time}. 
The accuracy of each component and more detailed settings of our profiling approach are available in Section~\ref{sec:eval-gimbal}. 
The online attack (Section \ref{sec:online_attack}) takes advantage of this constructed gimbal biasing pipeline to maximize the attack success rate.

\subsection{Online Attack}
\label{sec:online_attack}

As summarized above and illustrated in Figure~\ref{fig:attackdiag}, the online attack runs in two iterative loops: \emph{surrogate tracking} and \emph{planning–execution}. In this section, we first explain the underlying visual tracking vulnerabilities (Section~\ref{sec:tracking_vulnerability}) and key attack parameter choices (Section~\ref{sec:online_attack_para}), then detail the surrogate tracking loop and the planning–execution loop (Sections~\ref{sec:loop_surrogate}–\ref{sec:loop_planning}).

\subsubsection{Vulnerability of Visual Object Tracking}
\label{sec:tracking_vulnerability}

\begin{figure}[t]
\centering
\includegraphics[width=\linewidth]{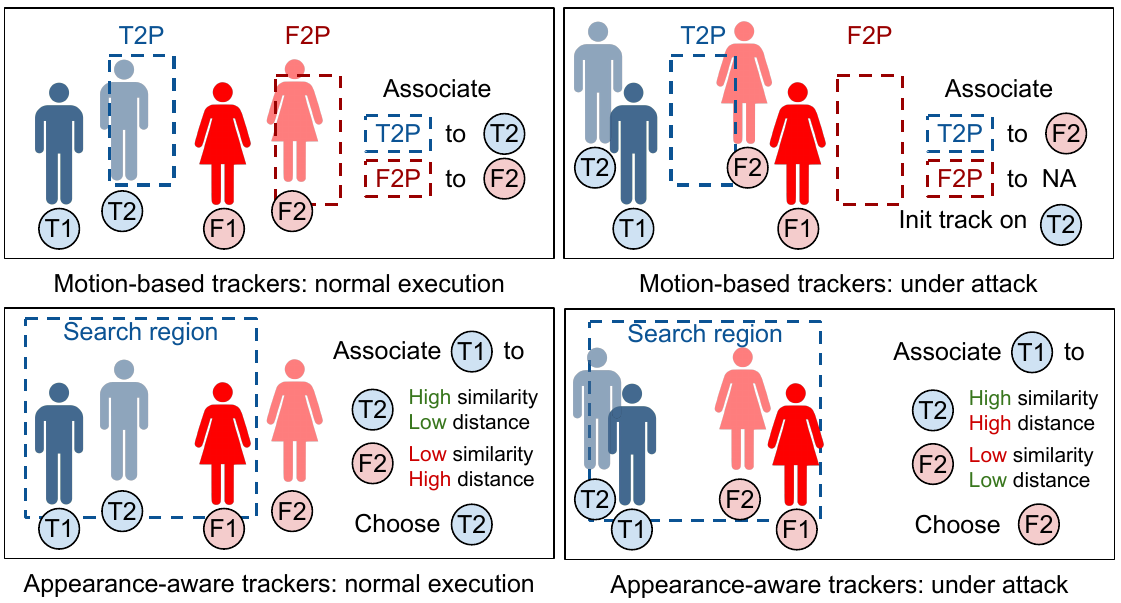}
\caption{Illustration of how visual tracking is compromised by malicious gimbal motion. The subfigures show the gimbal camera view; T/F stand for \sourceObj and \destinationObj; numbers 1/2 are frame IDs; P is the predicted object location in motion-based trackers. The attack triggers a yaw-axis gimbal rotation and all objects are shifted to the left.}
\label{fig:attack_insights}
\end{figure}

Visual object tracking processes a 2D image stream to localize a designated object across frames. While tolerant of occasional detection errors, trackers become unstable when gimbal motion is perturbed. Below, we explain the reasons, also illustrated in Figure~\ref{fig:attack_insights}.

\myparagraph{Motion-based trackers}
Such trackers (e.g., SORT~\cite{bewley2016simple}) rely on spatio-temporal consistency, such as bounding-box overlap and motion-model prediction, to associate objects across frames. These algorithms assume that target displacement between consecutive frames is small and predictable: they predict the object’s location using a motion model and then match the predicted box to actual detections based on distance or overlap. \rev{For instance, SORT uses a Kalman filter for motion prediction and Hungarian matching with IoU cost for object association.} Acoustic injection disrupts this assumption by inducing abrupt shifts in the gimbal’s field of view, creating large apparent jumps in target position. As a result, the tracker may fail to maintain association with the legitimate target and instead link the track to a nearby object. In Figure~\ref{fig:attack_insights}, the predicted location of \sourceObj (T2P) under attack falls closer to \destinationObj’s detection (F2), causing the tracker to switch its association.

\myparagraph{Appearance-aware trackers}
Appearance-aware trackers (e.g., Siamese trackers) employ deep similarity networks to match object patches across frames. However, they still depend on spatial proximity constraints to narrow the candidate search space. \rev{For instance, KCF~\cite{henriques2014high} considers object candidates only within a search region predicted from the last occurrence of the object; SiamRPN~\cite{li2018high} further penalizes candidates that are far from the search region center and whose scale deviates significantly from the previous frame.} Abrupt displacements caused by the adversarial gimbal motion could render \sourceObj falling outside or further from the search region, reducing the reliability of matching.

Therefore, in both categories, acoustic-induced gimbal perturbations could be optimized to undermine the fundamental assumptions of inter-frame correlation that tracking algorithms depend upon.

\subsubsection{Online Attack Parameters}
\label{sec:online_attack_para}

As introduced in the threat model (Section~\ref{sec:threat_model}), the attack seeks to mislead the tracker from following \textbf{\sourceObj} to \textbf{\destinationObj}. The attacker should determine the following parameters.

\myparagraph{Geometric relationship of the two objects}
Depending on the attack goal -- \emph{takeover} or \emph{escape}, the adversary can manipulate either \destinationObj or \sourceObj. By controlling the trajectory of one object, the attacker determines the relative positioning between the two, tailoring it to the vulnerabilities of the target gimbal system.
Specifically, in the offline profiling, we access which axis of the target gimbal is the most vulnerable to acoustic injection.
For example, if profiling reveals that the gimbal’s yaw axis is most susceptible to acoustic injection, the attacker should arrange \sourceObj and \destinationObj horizontally within the camera’s field of view. In this case, perturbations along the yaw axis can shift the tracker’s focus from \sourceObj to \destinationObj.
During the attack window, the two objects are typically placed in close proximity (e.g., 1–3 meters apart), as shorter separation reduces the required attack duration given the limited induced angular velocity.

\myparagraph{Cycle times}
The two in-loop processes should run as frequently as possible to enable fine-grained online optimization of the attack. However, several constraints govern the choice of cycle times. (1) Both loops must respect computational limits, which depend on the attacker’s hardware. In our case, the attack algorithm is lightweight, and its efficiency is evaluated in Section~\ref{sec:attack_efficiency}. (2) Stable gimbal motion requires that the planning–execution loop operate with a sufficiently long cycle time to allow acoustic signals to take effect. In practice, we align the planning-execution cycle time to an integer multiple of the attack signal frequency (Section~\ref{sec:gimbal-profile}). (3) The two loops must remain synchronized as the planning-execution takes the 3D object detection and surrogate model status as input. In practice, because the planning-execution loop is usually slower than surrogate tracking, its cycle time is set to be an integer multiple of the surrogate tracker’s cycle time, ensuring coordination.

\subsubsection{Loop of Surrogate Tracking}
\label{sec:loop_surrogate}

The attacker uses its own sensors to track objects in the scene, estimate the viewpoint of the target UAV gimbal camera, and run a black-box surrogate of the UAV's tracking algorithm that mimics its runtime behavior, preparing for later optimization.

\myparagraph{Real-time gimbal view estimation}
A key challenge of the black-box attack is that the video stream from the target gimbal is unavailable, limiting direct use of a surrogate tracker. To overcome this, \sys estimates the gimbal’s camera view in real time using external observations of the scene.
Specifically, the attacker first applies 3D object detection (via an external sensor suite) to obtain the world coordinates of \sourceObj, \destinationObj, and the UAV.
The gimbal’s orientation is then derived either by (1) a malicious IMU attached to the gimbal (requiring physical access), or (2) a high-resolution external camera with orientation detection algorithms (also see Section~\ref{sec:threat_model}). 
Additionally, camera parameters are already known from the offline profiling stage.
With the above information, \sys projects the 3D world coordinates of \sourceObj and \destinationObj into the estimated target UAV's 2D camera coordinate system. These projected 2D bounding boxes are then used as inputs to the surrogate tracking model, updated at each cycle of the loop.

\myparagraph{Surrogate tracking models}
Without access to the UAV system's image stream and internal data, we design surrogate tracking models to enable the optimization. 

\emph{Motion-based trackers}. 
Modern motion-based trackers typically operate in two steps: object detection followed by ID association.
In our surrogate model, object detection is replaced by the real-time gimbal view estimation. As such, at each frame, instead of running an image-based detector (which the attacker cannot access), we estimate the 2D bounding boxes of \sourceObj and \destinationObj and treat them as detection results. These estimated boxes are then the input of remaining surrogate tracking steps, with details depending on the specific surrogate model implementation. If using SORT~\cite{bewley2016simple} as the surrogate model, for instance, the two boxes are then matched with Kalman Filter (KF)~\cite{kalman1960new} predictions using the similarity metric of intersection over union (IoU), after which tracks are updated and KF states are refined.
To enable efficient optimization, we formulate this pipeline in a differentiable manner and compatible with gradient descent.

\emph{Appearance-aware trackers}.
Appearance-aware trackers localize targets by searching within a region around the last known position, leveraging both appearance cues and spatial consistency. In our black-box setting, the surrogate model cannot replicate the deep learning component that processes raw images. Instead, it directly estimates the deep learning component’s outputs and uses them as inputs to the subsequent object association stage.
Taking a SiamRPN-based surrogate model~\cite{li2018high} as an example, it crops a search region centered on the previous tracking box, typically four times its size. In the original tracker, this region is processed by a neural network to produce appearance-correlated proposals of boxes to be tracked. In our surrogate model, we mock this step by generating box proposals directly around the estimated positions of \sourceObj and \destinationObj in the gimbal camera view. Specifically, we sample multiple box proposals (e.g., 10) around each of the two objects, with slight random perturbations in their sizes and locations, and assign them high confidence scores (e.g., 1.0), reflecting the assumption that the component can reliably recognize foreground objects. These proposals are then passed to the association module, which selects the final tracked box. The entire process can also be implemented in a differentiable manner. Note that while the mocked proposals and confidence scores differ from those produced by the actual tracking with image feeds, they capture the key association behavior of appearance-aware trackers that they favor proposals closer to the search region center, which is sufficient for the optimization to exploit the tracker vulnerability.

\subsubsection{Loop of Planning and Execution}
\label{sec:loop_planning}

The attacker starts an iterative process with planning and execution run simultaneously in parallel when the surrogate tracking model is initialized and the targets are in position. The iterative process stops after the attack success or timeout.

\begin{algorithm}[t]
\caption{Planning of angular velocity to inject.}
\label{alg:angular_velocity_generator}
\small
\textbf{Input:} $S_i$: list of 3D bounding boxes for \sourceObj, \destinationObj, and UAV system at cycle $i$, 
$\mathbf{\theta}_i$: gimbal orientation ,
Surr: surrogate tracking algorithm ,
$\Delta P_i$: position offset to be injected ,
$t^c$: planning-execution cycle time. \\
\textbf{Output:} $\omega_{i+1}$: the planned angular velocity for cycle $i+1$.
\begin{algorithmic}[1]

\Function{FindOptimalAngularVelocity}{}
    \For{$iter =0$ to $N$}
        \State $T_{i:i+2}$ $\gets$ \Call{GenerateTrajectories}{$S_{i}$, $S_{i-1}$, $t^c$}
        \State $\Theta_{i:i+1}$ $\gets$ \Call{GenerateRotations}{$\mathbf{\theta}_{i}$, $\Delta P_i$, $t^c$}
        \State \Call{Update}{Surr, $T_{i:i+1}$, $\Theta_{i:i+1}$}
        \For{$tr$ in $T_{i+1:i+2}$}
            \State $X_{i+1}$ $\gets$ \Call{3DTo2DProject}{$tr$, $\omega_{i+1}$}
            \State \Call{ComputeLoss}{$X_{i+1}$, Surr}
        \EndFor
        \State \Call{GradientDescent}{$\omega_{i+1}$}
        \State \Call{ApplyConstraint}{$\omega_{i+1}$, $\Theta_{i}$}
    \EndFor
    \State \Return $\omega_{i+1}$ 
\EndFunction

\end{algorithmic}
\end{algorithm}

\myparagraph{Planning}
In the $i$-th planning-execution cycle, the algorithm computes the next attack plan $\Delta P_{i+1}$, i.e., the gimbal orientation offset to be executed in cycle $i+1$ (Algorithm~\ref{alg:angular_velocity_generator}). The algorithm first loads information from the surrogate tracking loop, including 3D detections of \sourceObj, \destinationObj, the UAV, and the latest surrogate model.  

To plan for the future, the algorithm predicts the states of objects, the gimbal, and the tracker. The attacker uses motion/trajectory prediction models to estimate the trajectories of \sourceObj, \destinationObj, and the UAV across cycles $i$ and $i+1$. Future gimbal motion is predicted using the current attack plan $\Delta P_i$ and offline profiling results. To increase robustness against prediction errors, we apply expectation-over-transformation (EoT): each predicted trajectory or motion is augmented with sampled noise, forming sets of trajectories for optimization. For each combination of sampled predicted states, the surrogate tracking model is executed on them and produces an estimated future tracker state, i.e., the location of the tracking box if the scene evolves according to the predicted trajectories.  
The optimization objective 
is then defined as the below:  
\begin{equation}
\label{eq:objective}
    \arg \min_{\omega<\omega^{max}} 
    \mathbb{E}_{P^t, P^f, X, G}
    \left[\sum_{i=1}^I \text{dist}(Box^{track}_i, Box^{false}_i)\right],
\end{equation}
where $P^t$, $P^f$, and $X$ denote the trajectory sets of \sourceObj, \destinationObj, and the UAV (from EoT), $G$ is the gimbal orientation distribution, $I$ is the number of frames in a cycle, $Box^{track}$ is the surrogate tracker output tracking box, and $Box^{false}$ is \destinationObj box. In short, the optimization searches for a gimbal angular velocity $\omega$ that maximizes the chance of misplacing the tracking box onto \destinationObj by minimizing the distance between the two boxes. The solution is constrained by the maximum achievable velocity $\omega^{max}$ from profiling. Because the surrogate tracking model is differentiable (Section~\ref{sec:loop_surrogate}), the optimization can be efficiently solved with gradient descent.

The final step of planning is to translate the angular velocity into the acoustic signal to inject. The translation is straightforward by applying the gimbal acoustic response model (Section~\ref{sec:gimbal_control}).

\myparagraph{Execution} 
The execution step runs in parallel with the planning step. The optimized attack plans (i.e., the acoustic signals to inject) are transmitted to the signal generator immediately after completion. When the $i$-th planning-execution cycle starts, the attack plan $\Delta P_{i}$ is immediately executed as it is generated in the last cycle (the first cycle has no execution step). This parallel scheduling minimizes the effect of optimization latency on attack performance, with efficiency shown in Section~\ref{sec:attack_efficiency}. 

%% file: sections/5_eval.tex
\section{Evaluation}
\label{sec:evaluation}
We extensively evaluate \sys to demonstrate its effectiveness and impact. 
First, we validate offline gimbal profiling, which underpins our runtime attack, by showing it consistently injects gimbal responses with directional bias (Section~\ref{sec:eval-gimbal}). 
Next, we evaluate attack effectiveness and robustness in high-fidelity simulation across diverse scenarios, including ablations (Section~\ref{sec:eval-sim}). 
Finally, we demonstrate practicality via physical experiments on a commercial drone, achieving results consistent with simulation (Section~\ref{sec:eval-phy}). 

\subsection{Gimbal Biasing Experiments}
\label{sec:eval-gimbal}

\begin{figure*}[t]
    \centering
    \begin{minipage}[t]{0.34\linewidth}
        \centering
        \includegraphics[width=\linewidth]{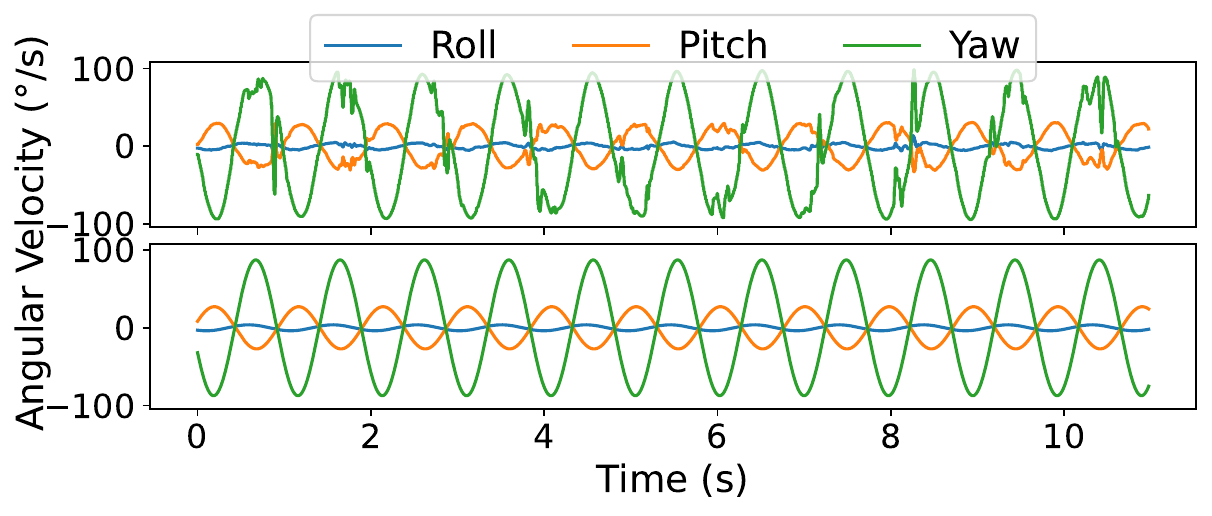}
        \caption{Accuracy of spectral analysis \revv{for 7744~Hz}. Reconstructed signal (bottom) is close to raw gyroscope readings (top).}
        \label{fig:gyrostack}
    \end{minipage}
    \hfill
    \begin{minipage}[t]{0.32\linewidth}
        \centering
        \includegraphics[width=\linewidth]{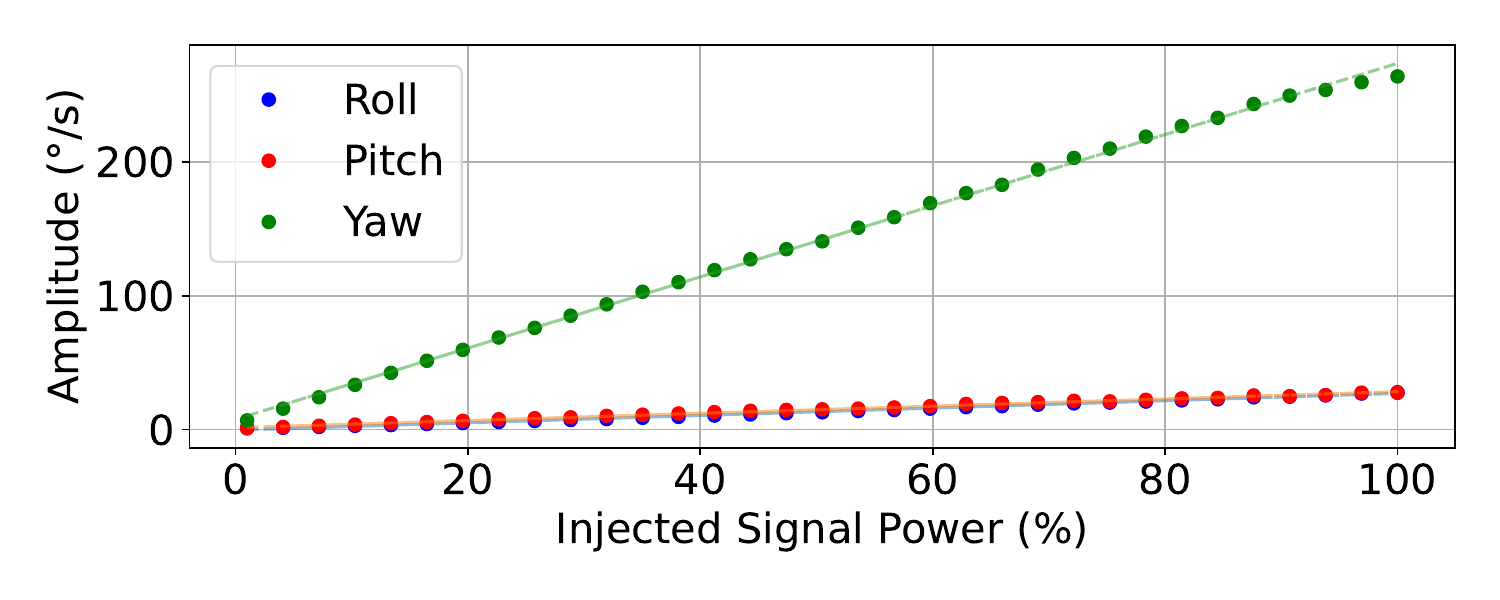}
        \caption{Accuracy of linear amplitude regression at 23232~Hz. }
        \label{fig:ampreg}
    \end{minipage}
    \hfill
    \begin{minipage}[t]{0.32\linewidth}
        \centering
        \includegraphics[width=\linewidth]{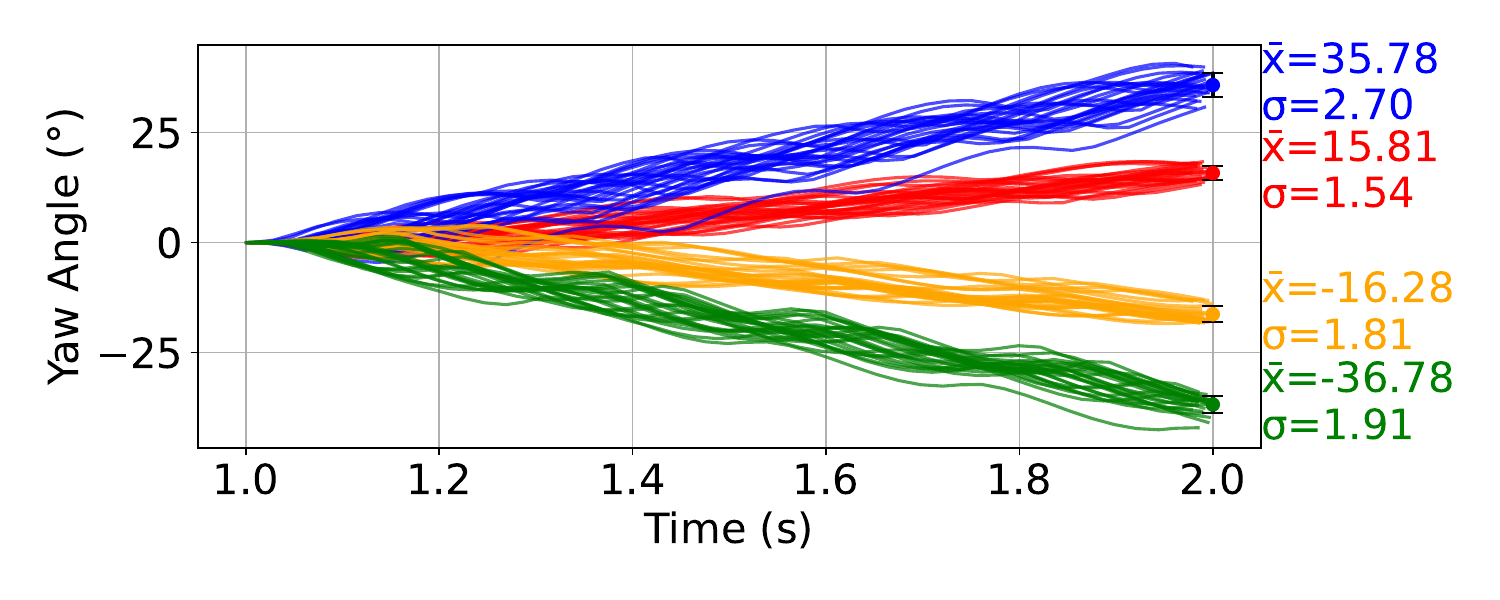}
        \caption{\rev{Online gimbal biasing: integrated angle over time for different amplitudes and directions.}}
        \label{fig:phaseanalysis}
    \end{minipage}
\end{figure*}

In this section, we evaluate the feasibility and consistency of offline gimbal profiling and biasing.

\myparagraph{Tested drone and gimbal systems}
We profile two state-of-the-art commercial UAV gimbal systems: a high-end industrial drone with a payload-mounted gimbal and a mid-end consumer drone with an integrated gimbal. For \droneModelOne, we perform full profiling and physical experiments, while for \droneModelTwo, we conduct frequency sweeps and directly adjust amplitudes in simulation-based evaluations.

\rev{\myparagraph{Experimental setup}
We attach a piezoelectric transducer directly to the gimbal housing and drive it with a function generator (AD9033) and amplifier (TDA8932) to inject crafted acoustic signals into the gimbal through vibration. An IMU (BMI160) is also attached to the gimbal housing to obtain the gimbal orientation.}


\myparagraph{Frequency sweep}
For both \droneModelOne and \droneModelTwo, we sweep 5 kHz to 30 kHz. After applying Least-Squares Spectral Analysis and local maxima filtering, we obtain resonant frequency sets \{7744, 23232\} and \{5526, 9214\} for \droneModelOne and \droneModelTwo respectively. \revv{Figure \ref{fig:gyrostack} shows a reconstructed signal using frequency, amplitude, and phase of oscillation in each axis obtained through spectral analysis.}

\myparagraph{Amplitude regression}
Figure~\ref{fig:ampreg} shows one regression produced at an injected signal frequency of 23232~Hz. We use a digitally controlled potentiometer to control the signal amplitude in terms of percentage power. 
We profiled 3 values \revv{$a_p=0.27, a_r=0.27, a_y=2.63$} for pitch, roll, and yaw, respectively. These three regressions achieve an average $R^2$ of $0.997$, showcasing the accuracy of our model.
\revv{Our measurements also show that, at this frequency, the motion response amplitude of the yaw axis is significantly more sensitive to acoustic signal strength than pitch and roll.}

\begin{figure*}[t]
    \centering
    \begin{minipage}[t]{0.32\linewidth}
        \centering
        \includegraphics[width=0.8\linewidth]{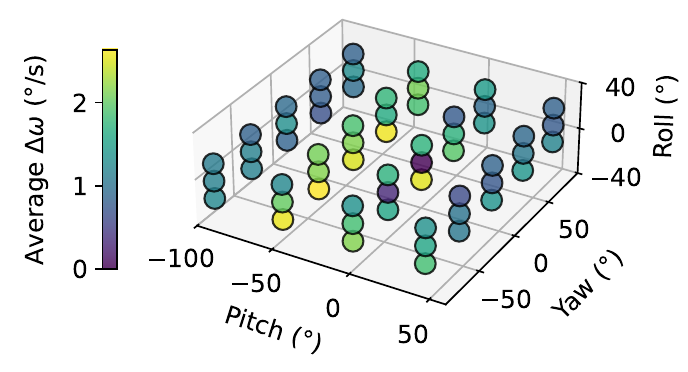}
        \caption{Orientation independence: difference from neutral orientation in amplitude, averaged across pitch/roll/yaw over various orientations.}
        \label{fig:orientationindependent}
    \end{minipage}
    \hfill
    \begin{minipage}[t]{0.32\linewidth}
        \centering
        \includegraphics[width=\linewidth]{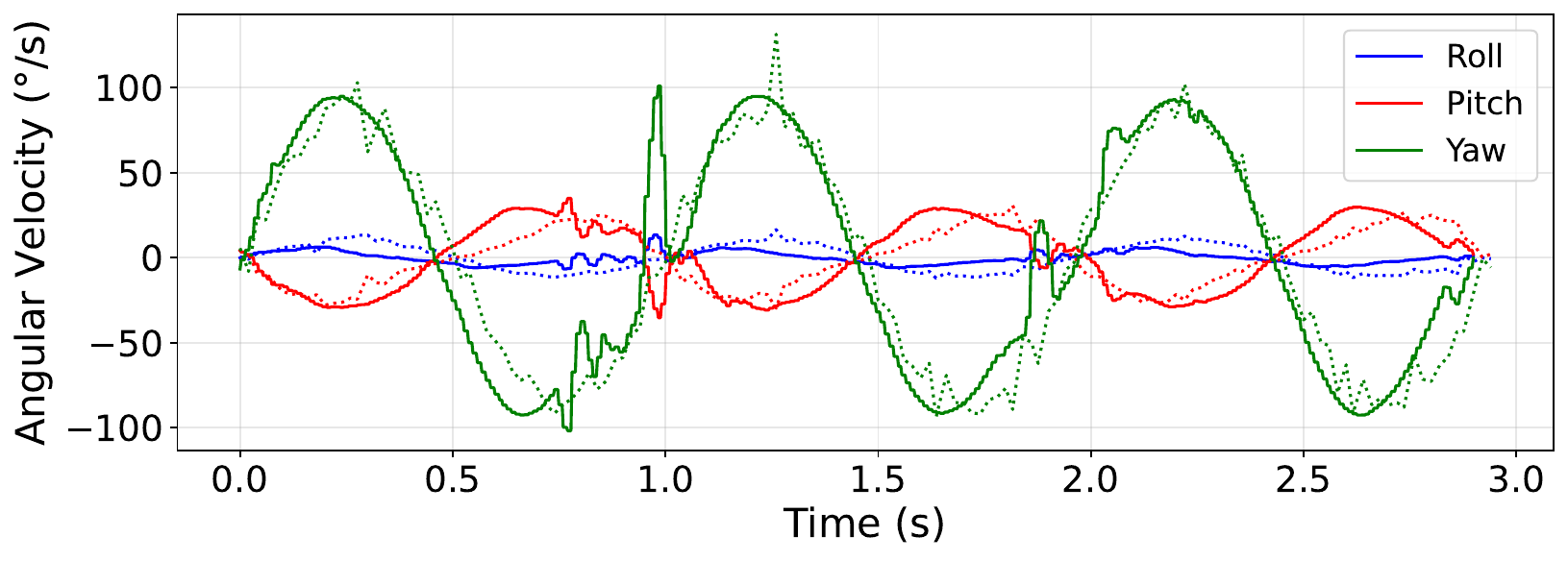}
        \caption{\revv{Acoustic gimbal motion injection trace in the physical world (solid line) and simulation (dotted line)}.}
        \label{fig:attack-realisticness}
    \end{minipage}
    \hfill
    \begin{minipage}[t]{0.32\linewidth}
        \centering
        \includegraphics[width=0.8\linewidth]{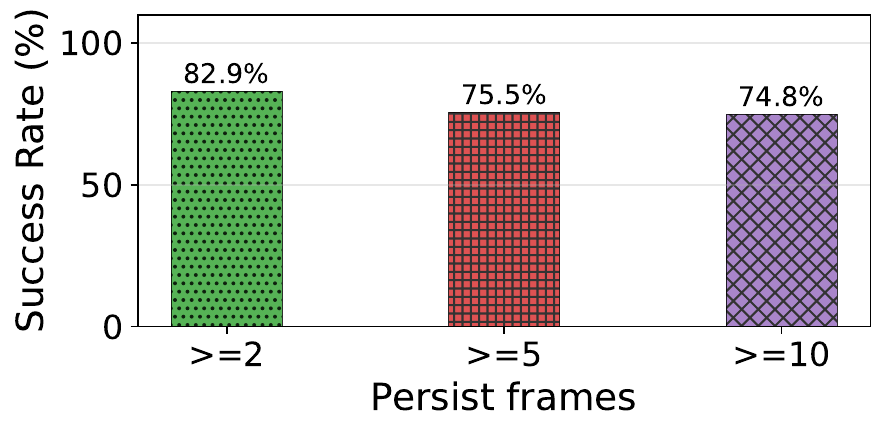}
        \caption{\rev{Distribution of number of consecutive frames of \switchAtk, showing that \switchAtk may become \lossAtk in attacks.}}
        \label{fig:persist_distribution}
    \end{minipage}
    
\end{figure*}

\myparagraph{Gimbal orientation independence}
To validate that acoustic gimbal biasing is independent of gimbal orientation, we test combinations of pitch at $-90^\circ$ to $45^\circ$, roll at $-15^\circ$ to $15^\circ$, and yaw at $-90^\circ$ to $90^\circ$, which covers the full range of possible orientations of \gimbalModel. In Figure~\ref{fig:orientationindependent} we plot the average error across all three axes at each orientation, compared to the motion observed at the neutral orientation. Our experiments show that across all orientations, the movement perceived by our external IMU is nearly identical with only slight variation. These minor errors can be most likely attributed to factors such as measurement error and any mechanical differences such as wear in the servos which rotate the different axes.

\myparagraph{Biasing results}
To evaluate the consistency and accuracy of our online biasing methodology, we conduct 100 trials at a frequency of 23232~Hz over 100\% and 50\% signal power in clockwise and counterclockwise yaw rotation. In Figure~\ref{fig:phaseanalysis} we plot all 100 traces integrated over a 1 second period. We successfully demonstrate that using our online biasing methodology, we can achieve consistent gimbal rotation in the desired direction with only a small degree of error.

\myparagraph{Execution latency} A potential concern is latency between acoustic injection and gimbal response. In our experiments, latency was negligible, consistent with expectations: the proof mass responds instantaneously, and gimbals must react in real time to support image stabilization.

\begin{figure}[t]
    \centering
    \includegraphics[width=\linewidth]{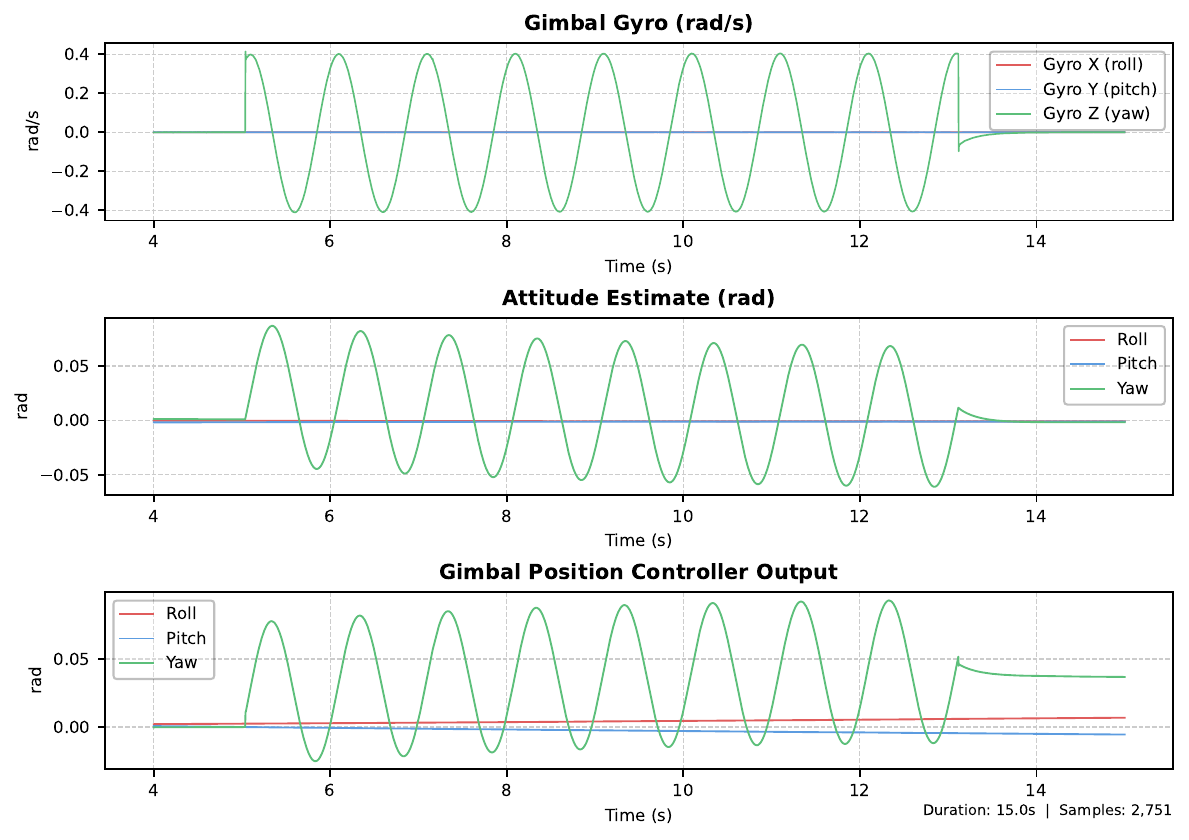}
    \caption{
    \rev{Time domain recording of intermediate gimbal states during acoustic injection.}
    }
    \label{fig:gimbal_log}
\end{figure}

\rev{\myparagraph{Time domain demonstration}}
\rev{To expose intermediate gimbal states and better understand acoustic injection, we simulate the gimbal stabilization pipeline in Gazebo~\cite{koenig2004gazebo}. Our implementation follows the STorM32-BGC design: IMU (gyroscope/accelerometer) measurements are fused via a complementary filter for attitude estimation, followed by a cascaded PID controller with an outer attitude loop and a high-gain inner rate loop. Acoustic perturbations are injected directly into gyroscope measurements (prior to Gazebo noise) as sinusoidal signals, with rectified variants for phase modulation along a selected axis.
Figure~\ref{fig:gimbal_log} shows gyroscope readings, attitude estimates, and controller outputs under a 1~Hz yaw-axis injection. The resulting sinusoidal motion is consistent with our empirical model despite pipeline nonlinearity.}

Overall, these results demonstrate both the feasibility of the profiling and the \rev{consistency of gimbal biasing} through acoustic injection.

\subsection{Simulation Experiments}
\label{sec:eval-sim}

Our simulation experiments comprehensively evaluate the attack’s effectiveness, robustness, and ablation results.

\begin{table}[t]
    \centering
    \renewcommand{\arraystretch}{0.6} 
    \setlength{\tabcolsep}{1.0pt} 
    \footnotesize
    \begin{tabular}{|c|c|c|c|c|c|c|}
        \toprule
        \textbf{Algorithm} & \textbf{Motion} & \textbf{Appear.} & \textbf{Robust Design} & \textbf{FPS} \\
        \midrule
        \textbf{KCF} \cite{henriques2014high} & \emptycircle & \fullcircle & - & 172\,/\,C \cite{henriques2014high} \\ 
        \textbf{SiamRPN} \cite{li2018high} & \emptycircle & \fullcircle & - & 49\,/\,G \cite{li2019siamrpn++} \\ 
        \textbf{DaSiamRPN} \cite{zhu2018distractor} & \emptycircle & \fullcircle & Distractor aware & 20.2\,/\,G\cite{fu2023siamese} \\ 
        \textbf{SORT} \cite{bewley2016simple} & \fullcircle & \emptycircle & - & 44\,/\,G\cite{ultralytics} \\ 
        \textbf{UCMCTrack} \cite{yi2024ucmctrack} & \fullcircle & \emptycircle & Motion compensate & 44\,/\,G\cite{ultralytics} \\ 
        \bottomrule
    \end{tabular}
    \caption{Details of evaluated tracking algorithms. Motion/appearance---tracker types; C/G---CPU or GPU.}
    \label{tab:algos}
\end{table}

\begin{table}[t]
        \centering
        \renewcommand{\arraystretch}{0.6}
        \setlength{\tabcolsep}{1.0pt}
        \small
        \begin{tabular}[t]{lcccc}
            \toprule
             & \rev{Normal} & \rev{Noisy} & \rev{\droneModelOne} & \rev{\droneModelTwo} \\
            \midrule
            \textbf{SORT}  & \rev{11.1\%/24\%} & \rev{0\%/75\%} & \rev{59.3\%/94.4\%} & \rev{55.6\%/96.3\%} \\ 
            \textbf{UCMC.} & \rev{0\%/0\%} & \rev{6\%/6\%} & \rev{68.5\%/92.6\%} & \rev{61.1\%/87.0\%} \\ 
            \textbf{Siam.} & \rev{0\%/0\%} & \rev{0\%/0\%} & \rev{83.3\%/88.9\%} & \rev{90.2\%/93.1\%} \\ 
            \textbf{DaSiam.} & \rev{0\%/0\%} & \rev{0\%/0\%} & \rev{92.6\%/98.1\%}  & \rev{97.2\%/100\%} \\ 
            \textbf{KCF}   & \rev{0\%/0\%} & \rev{17.6\%/17.6\%} & \rev{70.3\%/94.4\%}  & \rev{72.2\%/87.5\% }\\
            \bottomrule
        \end{tabular}
        \captionof{table}{Switch/disable attack success rates in simulation.}
        \label{tab:main_results}
\end{table}

\begin{figure}
    \centering
    \includegraphics[width=\linewidth]{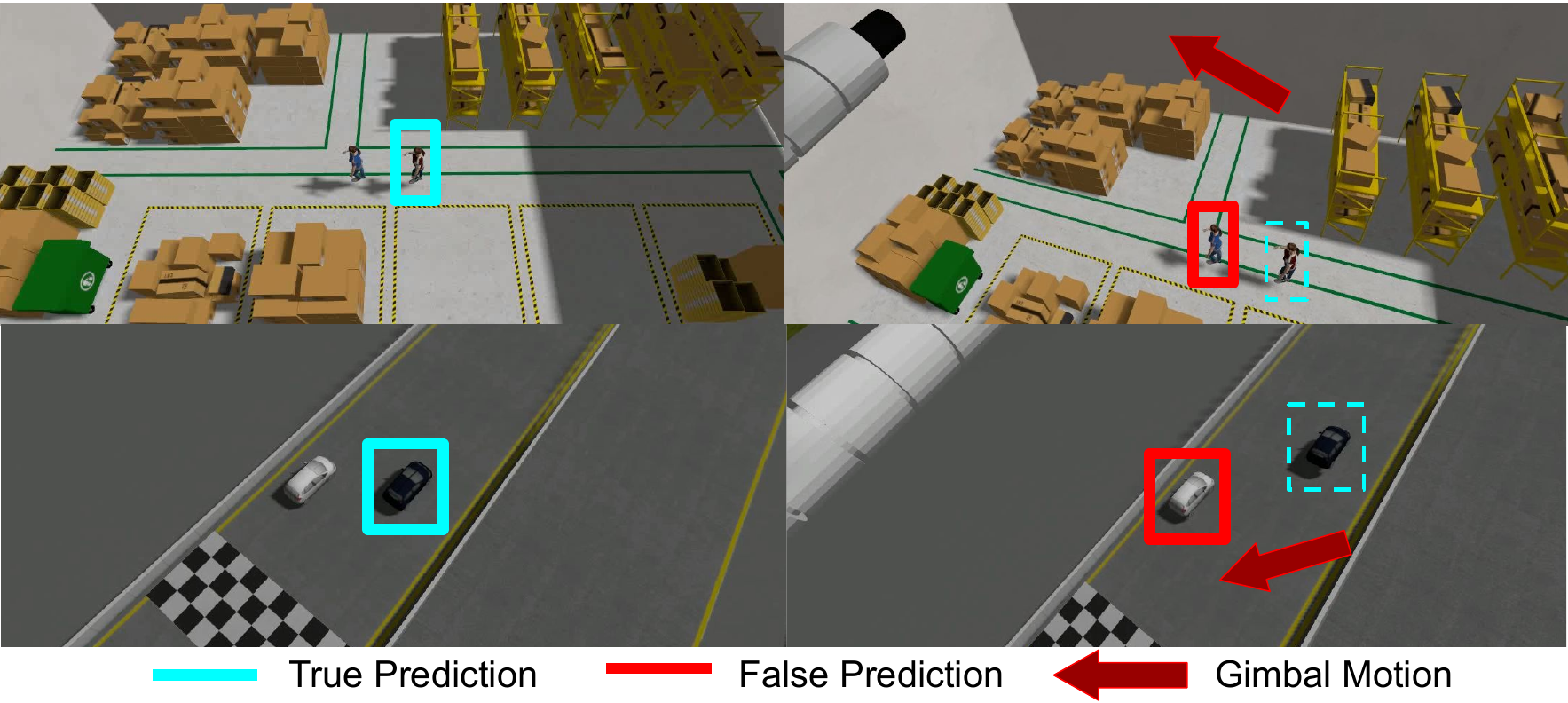}
    \caption{Demonstration of successful simulated \switchAtk attacks. }
    \label{fig:simulation-experiments}
\end{figure}

\subsubsection{Experiment Setup}
\label{sec:sim_exp_setup}
We begin by describing how the simulation is configured to approximate real-world attacks.

\myparagraph{Tracking algorithms}
We select five representative visual tracking algorithms, as shown in Table~\ref{tab:algos}. The algorithms span both major categories: motion-based trackers and appearance-aware trackers (details in Section~\ref{sec:tracking_vulnerability}).
In each category of tracker, we include the lightweight method (e.g., KCF and SORT), deep-learning-based methods (e.g., SORT and SiamRPN), as well as the state-of-the-art variants with robustness enhancement (e.g., DaSiamRPN and UCMCTrack).
For SORT and UCMCTrack which require an object detection component, we use the pretrained model YOLOv8x~\cite{ultralytics}.
For learning-based methods SiamRPN and DaSiamRPN, we use pretrained models trained on a range of datasets: VID \cite{russakovsky2015imagenet}, YoutubeBB \cite{real2017youtube}, COCO\cite{lin2014microsoft}, and ImageNetDet \cite{russakovsky2015imagenet}.
Our evaluated algorithms are lightweight enough to run on a typical UAV onboard computer. Other algorithms such as transformer-based methods are not suitable for real-time tracking on a mobile system (i.e. STARK~\cite{yan2021stark} runs at only a max 30-40 FPS even using a Tesla V100 GPU).

\myparagraph{Evaluation metrics} We evaluate the following metrics:
\begin{itemize}
    \item \textit{\SwitchAtk} is successful if the tracking box contains the center of \destinationObj for \rev{at least ten consecutive frames and persists to the end of the simulation.}
    \item \textit{\LossAtk} happens when the tracking box has no overlap with either \sourceObj or \destinationObj (often a side effect of failed \switchAtk).
    \item \textit{Disable}: The union of \switchAtk and \lossAtk attacks.
\end{itemize}

\myparagraph{Simulation scenario setup}
We simulate the attack with the high-fidelity physical simulator Gazebo~\cite{koenig2004gazebo} and the complete UAV software stack PX4-Autopilot~\cite{px4autopilot}. The simulation experiments are carried out on a desktop with Intel i9-14900K CPU and RTX 4080 GPU running Ubuntu 24.04. Figure~\ref{fig:simulation-experiments} demonstrates some of the attack scenes.

To demonstrate the generality of \sys, the test cases are distributed as follows: 
\begin{itemize}
    \item \textit{Worlds}: We include four distinct environments: factory, urban, field, and raceway. The factory world is an indoor setting, while the others are outdoor scenarios.
    \item \textit{Object categories, motion, and appearances}: Both \sourceObj and \destinationObj can be either pedestrians or vehicles. Pedestrians move at speeds of 0.5-1.5 m/s in random directions and vehicles move at speeds of 5-15 mph straight. Both pedestrian models (blue, green, red outfits) and vehicle models (silver, blue, and red hatchbacks) have pairwise distinct appearances.  
\end{itemize}
In total, we conduct 108 trials for each evaluation case. 54 are pedestrian cases, evenly distributed across three worlds (factory, urban, field) and six appearance pairs, with randomized initial motions. The remaining 54 trials involve vehicles, conducted in three worlds (raceway, urban, field), also covering six appearance pairs. 

\myparagraph{Simulation of the system and attacks}
For realistic simulation, we follow the threat model (Section~\ref{sec:threat_model}) and configure the simulation as follows:
\begin{itemize}
    \item \textit{Adversarial gimbal biasing}: We implement adversarial acoustic gimbal biasing in simulation using the gimbal acoustic response model (Section~\ref{sec:gimbal-profile}). 
    \rev{We inject Gaussian noise parameterized by the residual error from physical profiling via spectrum analysis (Section~\ref{sec:eval-gimbal}), modeling realistic motion perturbations (Figure~\ref{fig:attack-realisticness}).} 
    \rev{Specifically, we configure the gimbal maximum angular speed for the roll, pitch, and yaw gimbal axes as (0.04, 0.27, 2.35) and (0.76, 2.92, 0.21) rad/s, with realistic motion injection noise of (0.03, 0.04, 0.34) and (0.17, 0.51, 0.05) for \droneModelOne and \droneModelTwo respectively.}
    Other simulated gimbal characteristics, including rotation limits, degrees of freedom, and frame rate, are aligned with \droneModelOne or \droneModelTwo.
    
    \item \textit{UAV altitudes}: UAVs are above the ground by 10 meters and 25 meters for the pedestrian and vehicle scenarios, respectively, which aligns with real-world drone setup.
    \item \textit{Distance of two objects}: In the pedestrian scenario, \sourceObj and \destinationObj are placed 1–2 meters apart, while in the vehicle scenario they occupy adjacent lanes. Such distances are common in real-world settings and therefore do not appear suspicious as malicious behavior.
    \item \textit{Uncertain object motion}: 
    To reflect the real-world uncertainty, \sourceObj and \destinationObj move with injected Gaussian noise. Uncertainties in UAV flight are simulated with Gazebo's physical engine.
    \item \textit{Motion blur}: Abrupt camera shifts naturally introduce motion blur. We model this effect using linear and rotational blur kernels in OpenCV~\cite{opencv_library}. However, since the gimbal’s maximum angular speed is constrained, the blur remains minor, and our attack does not depend on it for success.
    \item \textit{Baselines}: \rev{In Normal, we simulate UAV object tracking without the presence of attack. In Noisy, we simulate UAV object tracking under strong and gusty wind supported by the realistic Gazebo WindEffects plugin.}
    \item The attack uses a surrogate model aligned with the evaluated tracker, with a 4~Hz planning–execution cycle, one gradient descent step per cycle, and three trajectory samples for expectation-over-transformation.
\end{itemize}

\subsubsection{Attack Effectiveness}

Table~\ref{tab:main_results} presents the attack results in the simulation setup as described in Section~\ref{sec:sim_exp_setup}. Our attack demonstrates overall success on different object tracking algorithms with \rev{\textbf{75.0\%}/\textbf{93.6\%}} averaged success rate for \switchAtk and disable (\switchAtk \& \lossAtk) respectively.
\rev{\LossAtk arises as a side effect of \switchAtk. Although the attack is optimized for \switchAtk, its probabilistic nature can still disrupt tracking and cause all objects to be lost. Figure~\ref{fig:persist_distribution} shows that 8.1\% achieve \switchAtk for at least two frames before transitioning to \lossAtk eventually.}

\begin{figure}[t]
    \centering
    \includegraphics[width=\linewidth]{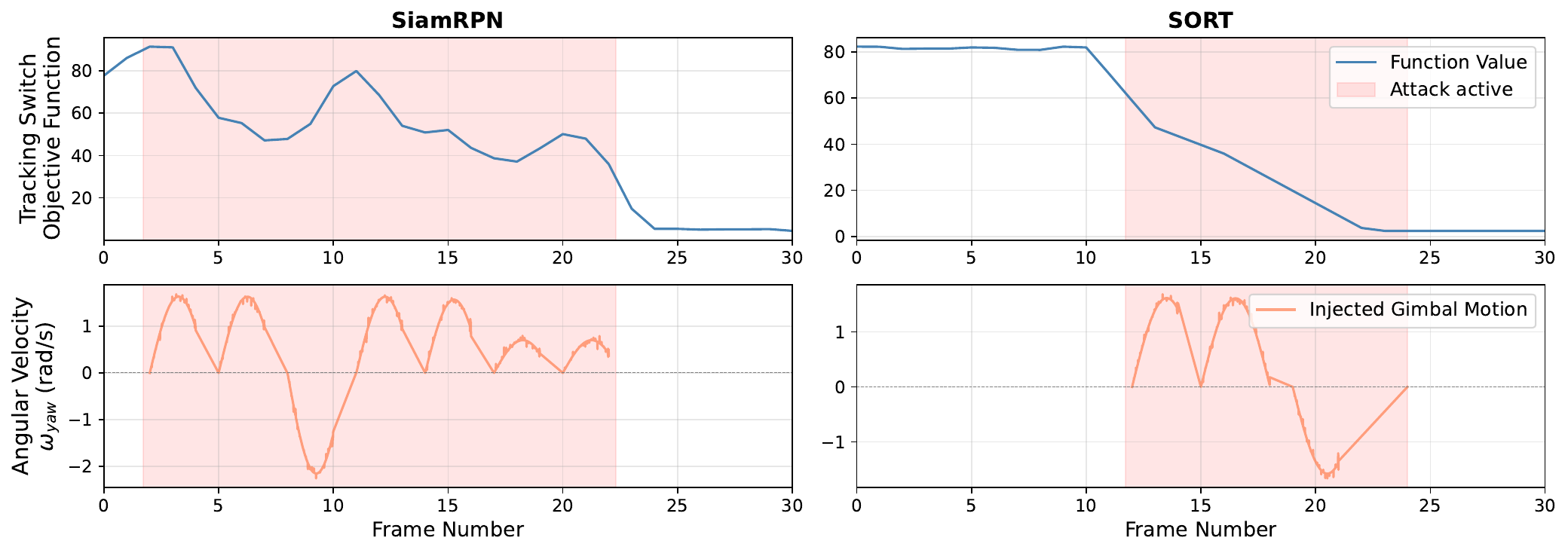}
    \caption{
    \rev{Time domain demonstration of injected gimbal motion induces target switch in simulation.}
    }
    \label{fig:time_domain_tracking_loss}
\end{figure}

\rev{\myparagraph{Time domain demonstration}}
\rev{Figure \ref{fig:time_domain_tracking_loss} shows a \switchAtk attack example for SiamRPN and SORT in simulation. The \switchAtk objective function (Equation \ref{eq:objective}) captures key attack success criteria. Minimizing the function value (blue curve) maximizes attack success. Orange curve shows simulated gimbal motion under acoustic injection attack, configured using the acoustic parameters obtained from the physical profiling experiment. It shows that injecting the optimized gimbal motion minimizes the objective function leading to \switchAtk attack success.}

\rev{\myparagraph{Baselines}
Switching is negligible under both settings, confirming that target switching does not arise in benign or naturally perturbed conditions. While noise can degrade tracking (e.g., causing \lossAtk for SORT), it does not induce sustained target switching. }

\myparagraph{Across tracking algorithms} Interestingly, more advanced tracking algorithms, i.e., with embedding of object appearance or compensation to camera instability, do not exhibit stronger resilience to our attacks.
Instead, appearance-aware trackers (SiamRPN, DaSiamRPN) achieve up to 29\% higher \switchAtk success rates than motion-based trackers (SORT, UCMCTrack).
This indicates that the attack perturbation is strong enough for corrupted spatio-temporal consistency to outweigh appearance features and the discriminative appearance awareness assists the attack by identifying the false object instead of losing in the background. The increasing \switchAtk success rate among KCF, SiamRPN, and DaSiamRPN with deeper appearance awareness provides further evidence. 

The advanced motion-based tracker UCMCTrack includes mechanisms to compensate for camera instability, yet it still fails under our \switchAtk attack, with a \switchAtk success rate higher than the vanilla SORT. The attack success DaSiamRPN and UCMCTrack  highlights that existing robustness designs cannot withstand the malicious perturbations but only assists the attack by reducing \lossAtk. \rev{Across three runs, the standard deviations of \switchAtk success (SiamRPN: $\pm 9.1\%$, DaSiamRPN: $\pm 3.5\%$, SORT: $\pm 4.5\%$, UCMCTrack: $\pm 8.2\%$, KCF: $\pm 10.3\%$) show the consistency of this trend.}
See Finding~\ref{finding:robust_trackers_fail}.

\begin{finding}[label=finding:robust_trackers_fail]
Robustness oriented tracking designs, including appearance aware and motion compensated trackers, do not withstand our \switchAtk attack; instead, they often increase \switchAtk success rates by preserving corrupted tracks rather than allowing \lossAtk.
\end{finding}

\myparagraph{Across different gimbals}. The attack demonstrates stable success across two commercial gimbal camera systems with distinct acoustic vulnerability shown in Section \ref{sec:eval-gimbal}. This suggests that the proposed online attack automatically adapts to specific gimbal parameters, regardless of the realistic random noise we added to the simulated gimbal motion.

\subsubsection{Analysis of Impacting Factors}

\begin{figure}[t]
    \centering
    \includegraphics[width=\linewidth]{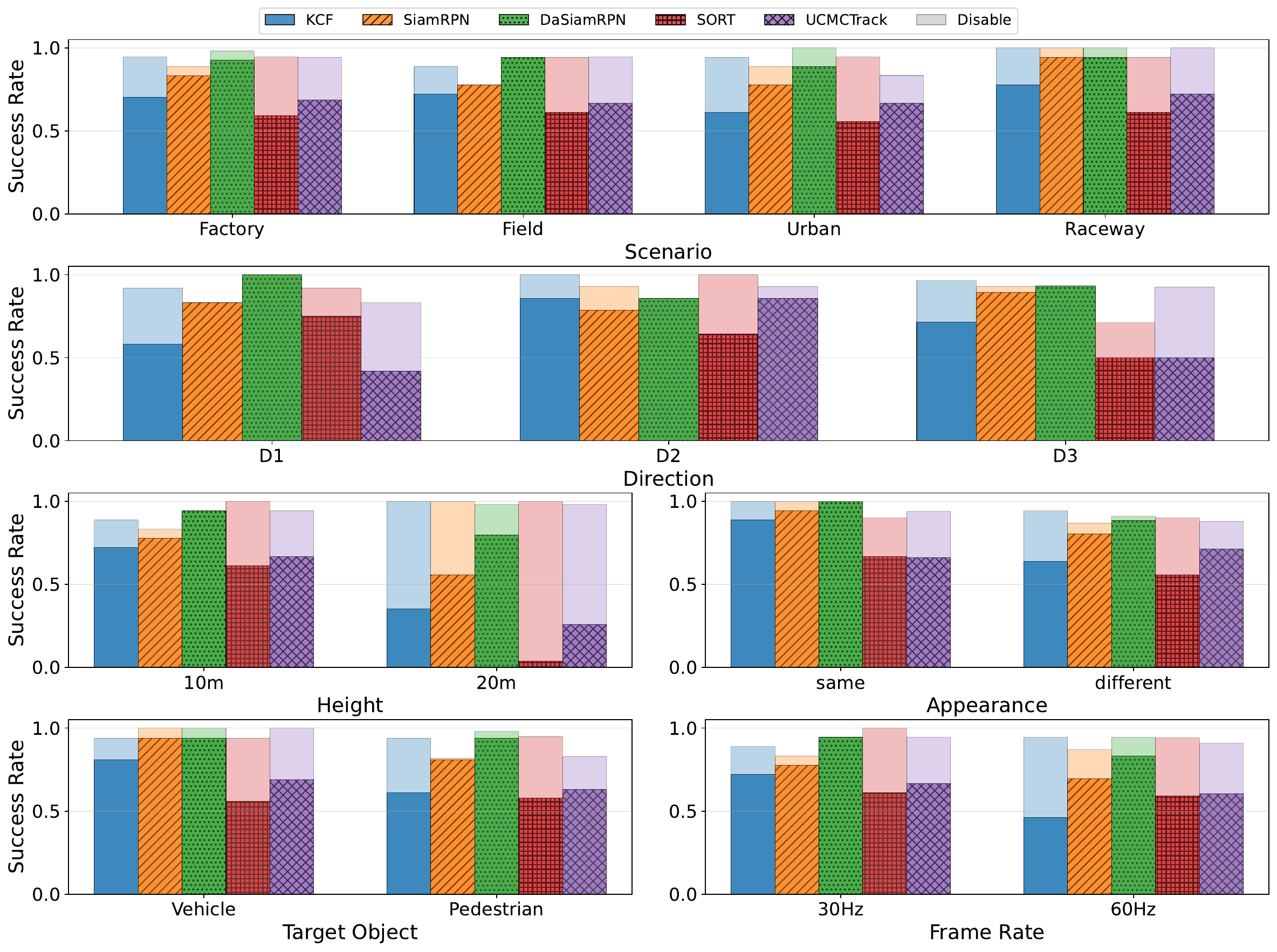}
    \caption{Simulation results analysis. Filled bars: \switchAtk success rate. Translucent bars: disable success rate.}
    \label{fig:asr_bars}
\end{figure}

Beyond the default simulation setup, we vary several factors reflecting real-world conditions to evaluate the robustness of the attack. Results are shown in Figure~\ref{fig:asr_bars}.

\myparagraph{Different environmental scenarios} Different environmental scenarios have only limited impact on \switchAtk success rates, showing a minimum of 70.2\% in the factory scenario and a maximum of 80.0\% in the raceway scenario. This is likely because the raceway vehicles are easier to track than the factory pedestrians because of object sizes.

\myparagraph{Different true target directions} This analysis focuses on the pedestrian scenario.
We categorize the randomized pedestrian motion into 3 directions in the gimbal camera field of view: D1 (moving away), D2 (moving horizontally), and D3 (moving closer). According to our results, different motion directions have limited impact on \switchAtk attack success, confirming that our optimization algorithm can handle variation in the \sourceObj's motion. 

\myparagraph{Different UAV heights}
The result shows attack success rates at varying UAV heights in the pedestrian scenario. The attack maintains high \switchAtk success at low altitude (10\,m), where richer visual cues and reliable detections benefit both category of trackers. However, \switchAtk success drops markedly at higher altitudes (20\,m) as smaller object size degrades tracking performance, leading to increased \lossAtk. The vehicle scenario results further support this hypothesis with a 78.9\% \switchAtk success rate at 30\,m.

\myparagraph{Target scenarios} 
The attack consistently achieves comparable or higher success rates in vehicle scenarios across all tracking algorithms. This indicates that targets which are inherently easier to track, such as vehicles with regular motion patterns and stronger visual features, are also more vulnerable to \switchAtk.

\myparagraph{Target appearance} 
Appearance-aware trackers show markedly higher \switchAtk success when the two objects appear identical, as expected from their reliance on appearance matching. These results indicate that an attacker can enhance success by intentionally making objects visually similar. 

\myparagraph{Camera frame rate} 
The 30 Hz setting, used by default in simulation, reflects common real-world configurations (Table~\ref{tab:algos}). Results show that 60~Hz frame rate mitigates the attack, as the same gimbal motion causes smaller inter-frame misalignment. Appearance-based trackers are most affected, suggesting that increasing frame rate can enhance robustness against malicious gimbal motion.

We conclude to Finding~\ref{finding:factors} summarizing the above:

\begin{finding}[label=finding:factors]
    The attack remains highly effective across real-world variations, and while factors such as higher UAV heights and higher tracking frame rates can reduce success, these challenging cases are limited.
\end{finding}

\begin{figure}[t]
    \centering
    \includegraphics[width=\linewidth]{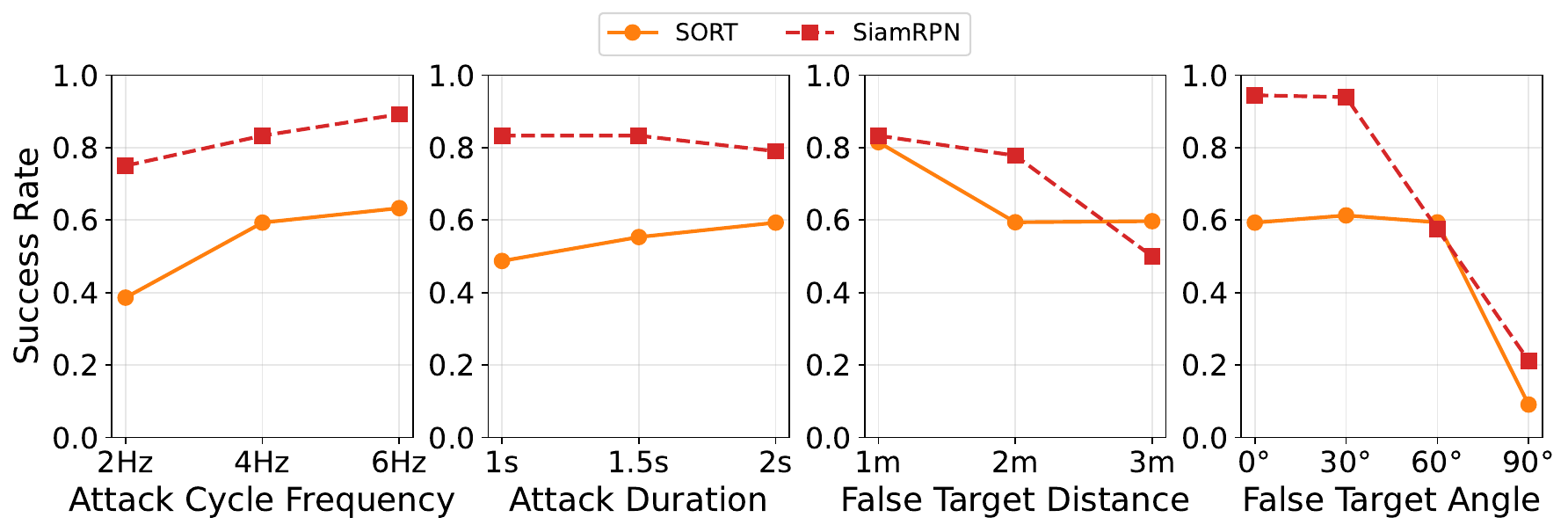}
    \caption{\SwitchAtk success rate w/ various conditions.}
    \label{fig:asr_dist_freq_max}
\end{figure}
\begin{figure}[t]
    \centering
    \includegraphics[width=\linewidth]{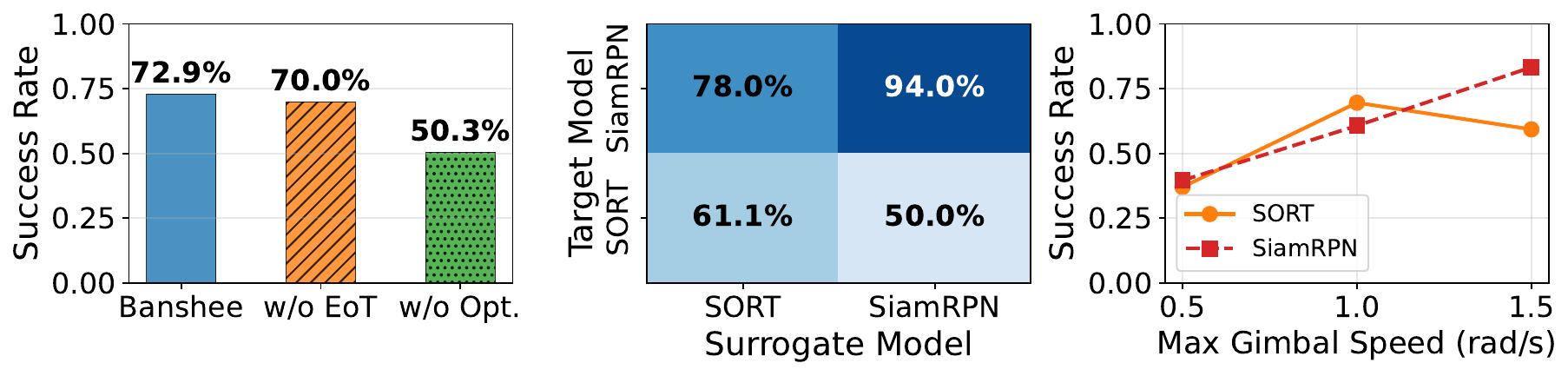}
    \caption{Left: attack components ablation. Middle: surrogate models transferability. Right: gimbal speed ablation.}
    \label{fig:ablation}
\end{figure}

\subsubsection{Ablation Study}
\label{sec:ablation}

We perform a set of experiments to understand the effect of different attack parameters and design components. Experiments are performed against SORT and SiamRPN with the same configuration as in Section~\ref{sec:sim_exp_setup} with \droneModelOne acoustic parameters.

\myparagraph{The usefulness of design components} 
We perform two ablation studies on the optimization component: removing expectation over transformation (EoT) and removing the entire optimization. Without EoT, optimization samples a single trajectory from the object motion models (\ref{sec:loop_planning}); without optimization, the attack applies a constant angular speed to shift the camera toward \destinationObj at the maximum controllable rate. As shown in Figure~\ref{fig:ablation}, both components substantially improve \switchAtk success.

\myparagraph{Distance between two objects}
In the pedestrian scenario, we evaluate attack performance as the distance between \sourceObj and \destinationObj varies (either may be the attacker controlled object; see Section~\ref{sec:threat_model} and Figure~\ref{fig:attack_insights}). For SiamRPN, \switchAtk success declines sharply with increasing distance, as its fixed search region limits association, making appearance-based trackers harder to compromise when objects are far apart. In contrast, motion-based trackers assign new IDs once geometric consistency breaks, enabling \switchAtk even at larger separations.

\myparagraph{Angle between two objects}
We vary the relative angle between \sourceObj and \destinationObj. At $0^\circ$ and $90^\circ$, the controlled object aligns with or is orthogonal to the gimbal axis most susceptible to acoustic injection (Section~\ref{sec:online_attack_para}). Deviations from this axis markedly reduce attack effectiveness (Figure~\ref{fig:asr_dist_freq_max}). The motion-based SORT is more sensitive to angle changes, as its association depends on whether the attacker lies along the injected motion direction, while the appearance-based SiamRPN is less affected due to its broader search region.

These results highlight that both distance and angle of the attacker-controlled object are critical. To maximize success, the attacker should align objects with the gimbal’s most vulnerable axis in the camera view—achievable through scene understanding and viewpoint prediction. The distance can then be tuned to balance success and stealth: greater separation slightly lowers \switchAtk probability but benefits inconspicuity. 
In summary:

\begin{finding}[label=finding:attacker_controlled_object]
    Attack success strongly depends on how the attacker places the controlled object, with favorable distance and alignment along the most vulnerable gimbal axis significantly boosting \switchAtk effectiveness.
\end{finding}

\myparagraph{Planning-execution cycle time}
A shorter cycle time yields more frequent optimization updates, making the attack more responsive to subtle scene changes. As shown in Figure~\ref{fig:asr_dist_freq_max}, shorter cycle time increases success rates, particularly for SORT. Ideally, the attacker would operate at the fastest cycle allowed by available computational power.

\myparagraph{Attack duration}
The impact of attack duration differs markedly between motion based and appearance aware trackers. 
For SiamRPN, high \switchAtk success is achieved with short (1\,s) perturbations, whereas the motion-based SORT tracker requires longer durations and additional online optimization to reach similar performance.
Empirically, excessively long attacks reduce stealth by causing noticeable camera shifts or leading to \lossAtk instead of \switchAtk. Hence, the attacker should determine an appropriate duration through offline testing on the target gimbal system.

\myparagraph{Maximum angular speed injected} 
Our attack remains stealthy, as the attacker-controlled object follows the victim at an unsuspicious distance (1--2\,m or in an adjacent lane) and the acoustic signal is invisible. On average, successful trials induce no more than a 30.2$^{\circ}$ gimbal movement. Moreover, tracker confidence levels remain consistent before and after the attack, suggesting that the perturbation is difficult to detect. Stealthiness can also be tuned by parameters such as maximum gimbal speed, enabling a trade-off between attack effectiveness and detectability. Figure~\ref{fig:ablation} shows attack success rates with respect to gimbal angular speed, and they are indeed strongly correlated.

Until now, we have identified multiple parameters that influence stealthiness, as summarized in Finding~\ref{finding:attack_stealthiness}.

\begin{finding}[label=finding:attack_stealthiness]
    An attacker can tune attack parameters to trade off stealthiness against effectiveness: placing objects too close makes their motion suspicious, while long attack durations or high gimbal angular speeds cause noticeable camera viewpoint changes.
\end{finding}

\myparagraph{Surrogate model choices} 
In Figure~\ref{fig:ablation}, we evaluate attacks using different combinations of surrogate and target trackers, such as applying SORT to attack SiamRPN and vice versa. Transferability is limited between motion-based and appearance-aware trackers, consistent with their distinct design principles. This suggests that attackers can perform offline attack before the online attack to select the surrogate model, or infer the type of the UAV tracker. See Finding~\ref{finding:attacker_knowledge}

\begin{finding}[label=finding:attacker_knowledge]
    Beyond gimbal profiling, the attacker can further improve success by inferring whether the UAV uses a motion-based or appearance-aware tracker and identifying effective attack durations or angular speeds.
\end{finding}

\rev{\myparagraph{Attacker-controlled object motion} 
Beyond noisy pedestrian motion, we evaluate more realistic motion models~\cite{moussaid2010walking, yang2020social}, where the attacker-controlled object follows physically plausible, smooth trajectories. Under the \droneModelOne main settings, the attack achieves 78.2\% (SiamRPN) and 67.2\% (SORT) \switchAtk success rates, demonstrating robustness to constrained and imprecise object motion.}

\rev{
\myparagraph{Gimbal motion pattern variations}
The injected motion pattern depends on resonant frequency and amplitude, which may drift due to environmental and hardware factors. To evaluate robustness, we execute an attack plan optimized for 4~Hz under perturbed conditions (3~Hz with profiled amplitudes and noise), resulting in a biased oscillation that differs from the planned pattern. The attack remains effective, achieving 77.8\% (SiamRPN) and 55.6\% (SORT) \switchAtk success.
}

\subsubsection{Attack Efficiency}
\label{sec:attack_efficiency}
Without dedicated optimization, our planning step runs at $\sim$7.6~Hz on our machine, which is fast enough to tolerate the selected cycle time (4~Hz). 
The main bottleneck for a real-time attack is the gradient optimization process. The attacker can parallelize the gradient optimization process (e.g., gradient step and EoT) using dedicated GPU devices.

\subsection{Physical Experiments}
\label{sec:eval-phy}

\begin{figure}
    \centering
    \includegraphics[width=\linewidth]{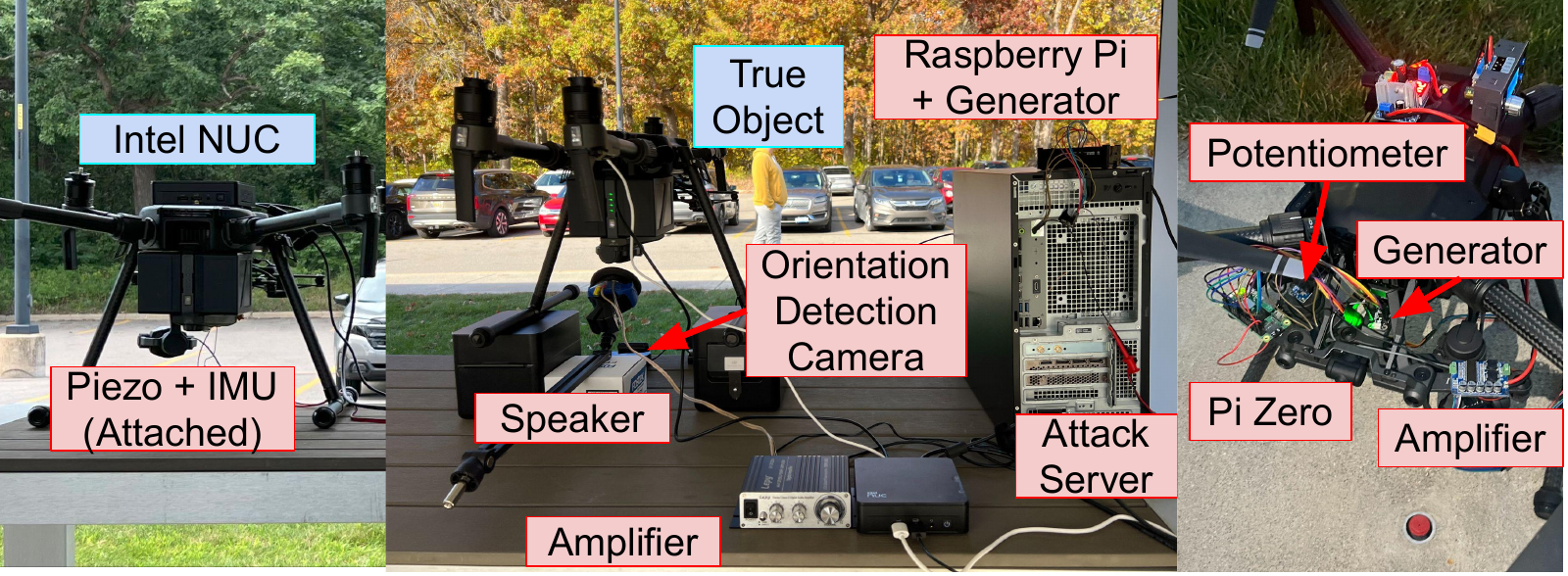}
    \caption{Benchtop direct-contact (left), benchtop contact-free (middle), in-flight direct-contact (right) attack testbeds.}
    \label{fig:benchtop-experiments}
\end{figure}

In this section, we demonstrate real-world attack performance through physical experiments on \droneModelOne.

\subsubsection{Experiment Setup}
As shown in Figure \ref{fig:benchtop-experiments}, we deploy three types of physical experiments.
\begin{itemize}
    \item \textit{Benchtop direct-contact}: The drone is placed on an elevated platform about 1.5 m above ground, representing typical consumer tracking scenarios at person height. 
    The gimbal camera is driven by an Intel NUC 13 Pro (Core i5), a common companion computer for onboard tracking algorithms.
    \rev{For acoustic injection, we attach a piezoelectric transducer as described in Section~\ref{sec:eval-gimbal}.}
    \item \textit{Benchtop contact-free}: 
    A Fostex FT17H speaker radiates the crafted acoustic signal through air. An EMEET C960 webcam observes the gimbal from below. The camera feed is processed with an optical-flow algorithm to estimate real-time gimbal orientation. Other hardware and configurations match the direct-contact benchtop setup. 
    \item \textit{In-flight direct-contact}: 
    We build a compact malicious payload with multiple low-power integrated-circuits (which could be reduced to a single embedded unit with further engineering), containing a Raspberry Pi Zero 2W, function generator (AD9033), amplifier (TDA8932), digital potentiometer (DS3502), and IMU (BMI160). The Pi Zero controls the generator and potentiometer to produce the crafted waveform, delivered by a piezoelectric transducer mounted on the gimbal housing. During tests, the drone hovers at about 5 meters. 
\end{itemize}

The online attack algorithm runs on a desktop with Intel i9-14900K CPU and RTX 4080 GPU. An EMEET C960 Webcam is used to obtain the input for the surrogate tracking algorithm. \rev{We use an appearance-aware surrogate tracker as it outperforms motion-based alternatives on \droneModelOne, indicating the tested drone may utilize appearance-aware visual tracking.}

We evaluate five representative tracking algorithms (Table~\ref{tab:algos}) in benchtop attacks, and the \droneModelOne built-in \trackAlgorithm in the in-flight setting. All trackers achieve real-time performance ($\sim$30 FPS or higher) on an Intel NUC 13 Pro (Core i5), except DaSiamRPN. For each tracker, we conduct 10 trials with two pedestrians of distinct appearance and a false-target separation of approximately 3\,m. The attack uses a 4~Hz planning–execution cycle, with one gradient descent step per cycle and three trajectory samples for expectation-over-transformation.

\begin{table}[t]
    \centering
    \renewcommand{\arraystretch}{0.6} 
    \setlength{\tabcolsep}{1.0pt} 
    \small
    \begin{tabular}{lcccccc}
        \toprule
         & FPS & Direct-contact & Contact-free & In-flight \\
        \midrule
        \textbf{SORT} & 25.0 & 60\%/100\% & 60\%/100\% & - \\ 
        \textbf{UCMCTrack} & 24.9 & 80\%/100\% & 100\%/100\% & - \\ 
        \textbf{SiamRPN} & 23.4 & 90\%/100\% & 90\%/90\% & - \\ 
        \textbf{DaSiamRPN} & 7.5 & 90\%/90\% & 100\%/100\% & - \\ 
        \textbf{KCF} & 30.0 & 70\%/100\% & 70\%/90\% & - \\
        \textbf{\droneModelOne} & - & - & - & 60\%/80\% \\
        \bottomrule
    \end{tabular}
    \caption{Target switch/disable attack success rate of real-world physical experiments.}
    \label{tab:physical}
\end{table}

\subsubsection{Physical Experiment Results} The attack achieves an average \switchAtk success rate of \textbf{79.1\%/95.5\%} across all benchtop and in-flight trials (Table~\ref{tab:physical}). The overall performance is consistent with the simulation results (Table~\ref{tab:main_results}), validating the attack’s practicality in the real world.

Across the three experiment settings, as the scenario becomes more challenging in terms of reduced attacker capability and increasing real-world uncertainty and latency, the proposed attack demonstrates stable success. 
The comparable results between benchtop direct-contact and benchtop contact-free attacks demonstrate that physical access to the victim UAV is not necessary to launch the attack.

\begin{figure}[t]
    \centering
    \includegraphics[width=1.0\linewidth]{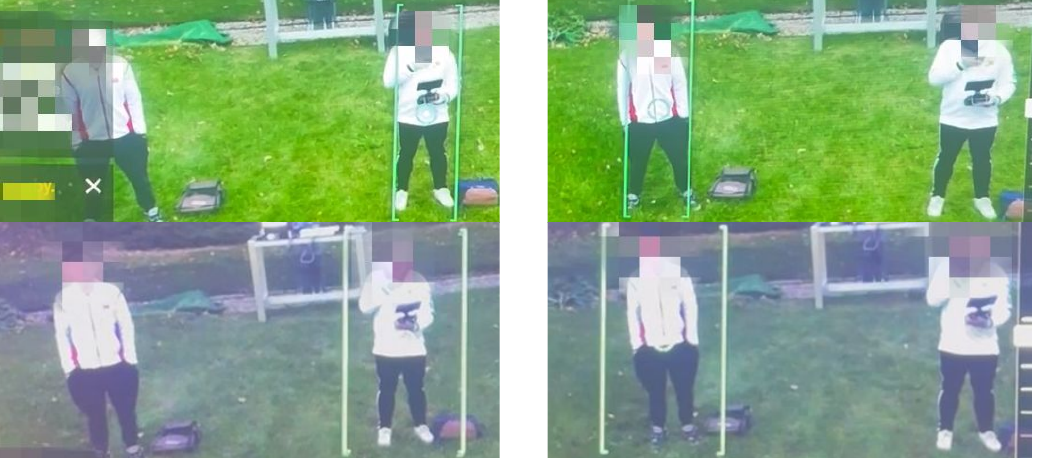}
    \caption{Before/after \switchAtk \trackAlgorithm. First row: daytime. Second row: at dusk.}
    \label{fig:inflight_attack}
\end{figure}

More importantly, the attack achieves 60\% \switchAtk success rate against the built-in tracking of the industrial-grade \droneModelOne during flight, with complete black-box knowledge (Figure \ref{fig:inflight_attack}). This not only demonstrates that the attack is physically achievable against an operating UAV system but also confirms this security vulnerability in commercial closed-source UAVs.

Failure cases of \switchAtk (\lossAtk and tracking maintained) typically include one or more transient \switchAtk successes across non-consecutive frames. This suggests that the failures arise from acoustic control noise in real-world environments rather than inherent robustness in the target system.
The results lead to Finding~\ref{finding:physical_experiments}.

\begin{finding}[label=finding:physical_experiments]
    Physical experiments confirm that both direct-contact and contact-free attack vectors can corrupt UAV tracking, including successful exploits on a fully closed-source drone, although real-world environmental noise introduces additional challenges for the acoustic attack.
\end{finding}


%% file: sections/6_disc.tex

\section{Discussion}
\label{sec:discussion}

\myparagraph{Potential defenses (details in Appendix~\ref{sec:app_defenses})}
Several mitigation strategies can be applied against \sys, but each has notable limitations. Hardware approaches aim to eliminate the attack vector via acoustic isolation of the gyroscope~\cite{ji2021poltergeist, trippel2017walnut} or improved signal conditioning and ADC design~\cite{trippel2017walnut}. However, these solutions are costly, require modifications to off-the-shelf components, and often entail substantial gimbal redesign, making them primarily viable for future UAV systems. Software defenses are easier to deploy, but remain challenging: \sys induces large, rapid motions, while image stabilization demands high-rate processing. Existing detection methods based on simple motion models~\cite{quinonez2020savior, wang2025vimu, meng2025mars}, sensor fusion~\cite{lee2025gyromag}, or specialized hardware~\cite{wang2024adcbank} also do not align well with typical UAV gimbal architectures.

There are promising defense directions. One option, inspired by GPS spoofing defenses, is to use image-based signals such as visual odometry to detect inconsistencies between visual and inertial measurements~\cite{gu2021coupleddetection, varshosaz2020civilian}. This approach is lightweight and software-based, but can be bypassed by sophisticated adversaries, as visual pipelines themselves are also vulnerable~\cite{davidson2016controlling, ranjan2019atkoptflow, schmalfuss2022perturbflow}.

\myparagraph{\rev{Attack controllability}}
\rev{Our attack achieves empirical success using a learning-based gimbal response model under a simplified control abstraction. A more principled control theory analysis of modern gimbals with multi-loop control and sensor fusion could further improve the fidelity of this gimbal control approximation. In addition, reliably selecting a specific target in the presence of multiple nearby objects remains an open challenge for practical deployment. We leave these directions to future work.}

\rev{
\myparagraph{Remote attack feasibility}
While we demonstrate attack success in physical experiments under the benchtop contact-free setup, we observe that motion injection is sensitive to sound pressure level (SPL). In our setup, placing the speaker approximately 4 inches below the gimbal achieves $\sim$110\,dB, inducing sufficient motion. Prior work~\cite{son2015rocking, trippel2017walnut, tu2018injected, ji2021poltergeist, gao2022kite, 285403} demonstrates acoustic attacks over distances ranging from 10\,cm to 7.6\,m, while long-range ultrasonic emitters~\cite{ji2021poltergeist, longRangeUltra} and laser-based approaches~\cite{shamsiwip} can potentially extend attack distances beyond 100\,m.}

\myparagraph{Threats to validity}
While simulation enables diverse scenarios, it cannot fully capture real-world complexity. Our physical experiments validate the attack’s practicality in physical world, though under limited scenario diversity. Future work will expand real-world evaluations to additional gimbal and system models.



%% file: sections/7_conc.tex
\section{Conclusion}
This paper introduces \sys, the first physically realizable \switchAtk attack on UAV visual tracking via acoustic injection on gimbal-camera systems. By empirically profiling gimbal acoustic responses, \sys generates perturbations that induce directionally biased motion and probabilistically redirect tracking to another object. Our pipeline achieves $>90$\% success in simulation and successful black-box attacks on a commercial drone. These results show that acoustic vulnerabilities extend beyond sensor disruption to application-level compromise.

\myparagraph{Acknowledgments} 
We thank the anonymous reviewers and our shepherd for their valuable feedback and constructive suggestions throughout the revision process.
This work was partly supported by NSF under the Grant CNS-2321532 and the National AI Institute for Edge Computing Leveraging Next Generation Wireless Networks, Grant \# 2112562.

%% file: sections/8_ethi.tex
\section*{Ethical Considerations}

This work reveals an end-to-end vulnerability in UAV visual tracking systems, linking acoustic injection at the sensor level to application-level target-following failures. Given the potential safety risks, we take several measures to ensure responsible conduct.

\myparagraph{Responsible disclosure}  
We have contacted the vendor of the UAV and gimbal systems used and shared our findings prior to publication. Product identifiers are anonymized to reduce misuse risk. We will follow coordinated disclosure practices and release full details only after mitigations are available.

\myparagraph{Controlled experiments}  
All physical experiments were conducted in controlled environments using a single UAV at low altitude within confined areas. Tests were performed under supervision, including benchtop setups and carefully managed outdoor trials, ensuring no risk to bystanders, property, or other aircraft.

\myparagraph{Limited artifact release}  
To support reproducibility, we release simulation code, profiling tools, and non-sensitive datasets, but exclude any software or hardware details that enable direct acoustic exploitation. The artifacts allow validation in simulation and study the implications for defense without providing end-to-end attack capabilities.

\myparagraph{Purpose of research}  
Our goal is to expose a critical class of vulnerabilities in widely deployed UAV systems, not to enable misuse. By demonstrating the feasibility of acoustic-induced target switching, we highlight the need for more robust gimbal design, sensor fusion, and tracking algorithms. We believe responsible disclosure of these findings will support timely mitigation by manufacturers, researchers, and regulators.

In summary, this work follows principles of responsible disclosure, controlled experimentation, limited release, and a focus on improving UAV system safety and robustness.


\section*{LLM Usage Considerations}

\myparagraph{Originality}
LLMs were used only for editorial refinement (e.g., clarity, grammar, and style). All scientific content—including problem formulation, methodology, experiments, and analysis—was developed and validated by the authors. The literature review and citation decisions were performed manually. In accordance with conference guidelines: \emph{``LLMs were used for editorial purposes in this manuscript, and all outputs were inspected by the authors to ensure accuracy and originality.''}

\myparagraph{Transparency}
LLMs are not part of our methodology or evaluation. All experiments and analyses are fully reproducible without any LLM service, and no technical contribution or conclusion depends on LLM outputs.

\myparagraph{Responsibility}
We did not train or fine-tune any LLMs or collect data for that purpose, nor did we provide proprietary or sensitive information to LLM services. As such, the environmental and ethical impact is minimal, and all contributions remain the authors’ own.

%% file: sections/9_appe.tex
\appendices

\section{Detailed Discussion of Defenses}
\label{sec:app_defenses}

\myparagraph{Physical layer}
A direct mitigation is to eliminate the attack vector via improved physical housing of the gimbal gyroscope. Acoustic isolation using microfibrous metallic fabric, MEMS acoustic metamaterials, or acoustic foam can attenuate injected signals~\cite{ji2021poltergeist, trippel2017walnut}. However, these approaches increase weight, cost, and design complexity, and may still be bypassed by strong acoustic signals.

\begin{figure}[h]
    \centering
    \includegraphics[width=0.7\linewidth]{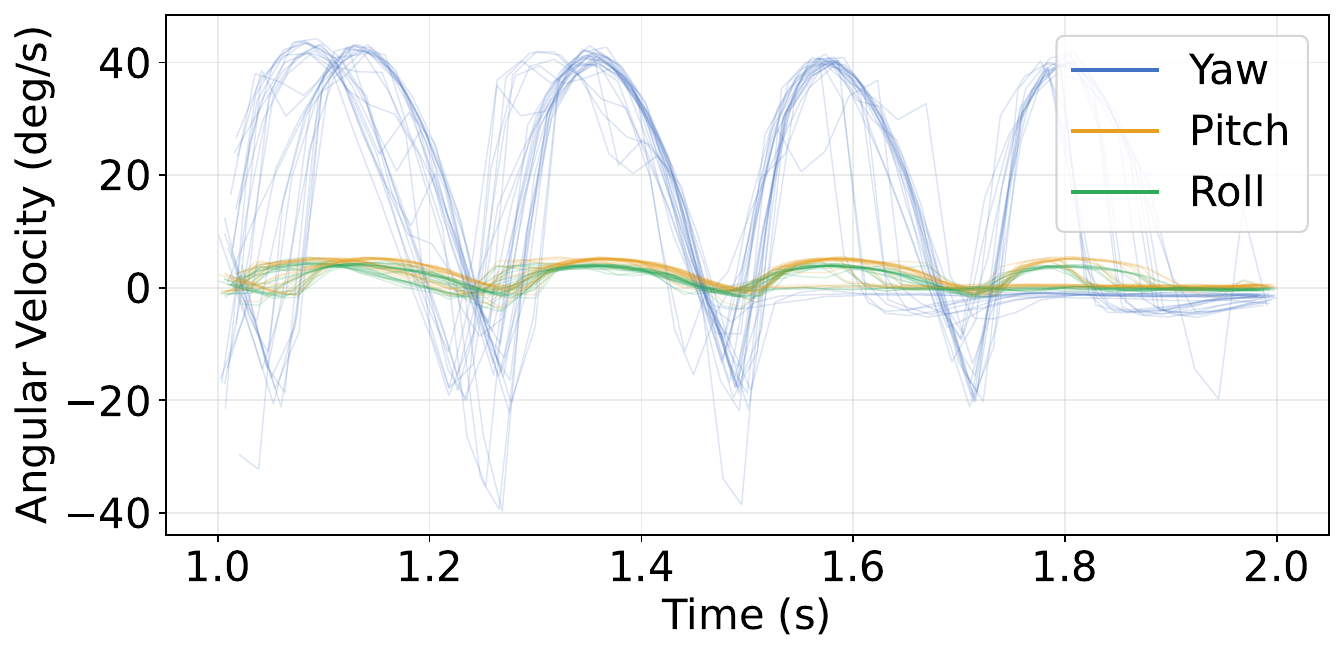}
    \caption{\rev{3-axis angular velocity readings over 25 consecutive injections with directional biasing.}}
    \label{fig:velocityrepeated}
\end{figure}

\myparagraph{Hardware layer}
Secure sensor design can mitigate acoustic injection by strengthening signal conditioning in MEMS IMUs. Gyroscopes and accelerometers infer motion from small proof-mass voltages that are amplified, filtered, and sampled. Acoustic attacks exploit weaknesses in this pipeline, such as amplifier clipping (introducing DC bias) or inadequate low-pass filtering. Secure sampling techniques, e.g., randomized or out-of-phase sampling aligned with sensor resonance, can further reduce vulnerability~\cite{trippel2017walnut}. However, these defenses require IMU or ADC modifications and are typically infeasible for commodity UAV systems.

\myparagraph{Software layer} 
Software-based defenses include digital image stabilization and learning-based denoising~\cite{ji2021poltergeist, jeong2023unrocking}. However, \sys induces much larger motion than typical blur, limiting the effectiveness of stabilization. Learning-based filtering can suppress injected noise but requires large models and extensive training, introducing latency incompatible with real-time gimbal stabilization.

\myparagraph{Detection}
Detection methods include model-based approaches (e.g., SAVIOR~\cite{quinonez2020savior}, VIMU~\cite{wang2025vimu}, MARS~\cite{meng2025mars}) that predict IMU readings, and sensor-fusion approaches (e.g., Gyro-Mag~\cite{lee2025gyromag}, ADC-Bank~\cite{wang2024adcbank}) that detect inconsistencies across sensors. However, model-based methods assume tractable system dynamics (e.g., rigid-body or aerodynamic models), which do not apply to gimbals with nonlinear controllers and friction. Sensor-fusion methods require additional sensors or hardware redundancy, which are often unavailable in gimbal cameras and may themselves be vulnerable to attacks~\cite{sathaye2022experimental, kune2013ghost}.

\myparagraph{Camera motion compensation}
Incorporating camera motion compensation (CMC) into tracking can reduce attack success rates, especially when trading \switchAtk for \lossAtk. However, the reduction is insufficient to fully mitigate \sys. Moreover, CMC incurs significant computational overhead, limiting feasibility on resource-constrained UAVs, particularly when combined with heavy tracking models~\cite{yi2024ucmctrack}.

\myparagraph{Visual odometry}
Visual odometry (VO), which estimates camera egomotion from image sequences, can be used to detect acoustic injection by comparing VO estimates with gyroscope readings. Under normal operation, the two are aligned; under attack, they diverge. VO can run in parallel with stabilization and tracking, requiring only the onboard camera, making it lightweight and deployable via software updates. However, VO is also vulnerable to spoofing, and a sophisticated attacker may evade detection by aligning both modalities~\cite{davidson2016controlling, ranjan2019atkoptflow, schmalfuss2022perturbflow}.

\myparagraph{Recovery}
Upon detection, the system can enter a recovery mode, e.g., limiting gimbal angular velocity to reduce attack impact, which we observe lowers attack success rates. Alternatively, the UAV can hover or land to preserve safety. These strategies improve robustness but may degrade normal tracking performance.

\begin{figure}[t]
    \centering
    \includegraphics[width=0.8\linewidth]{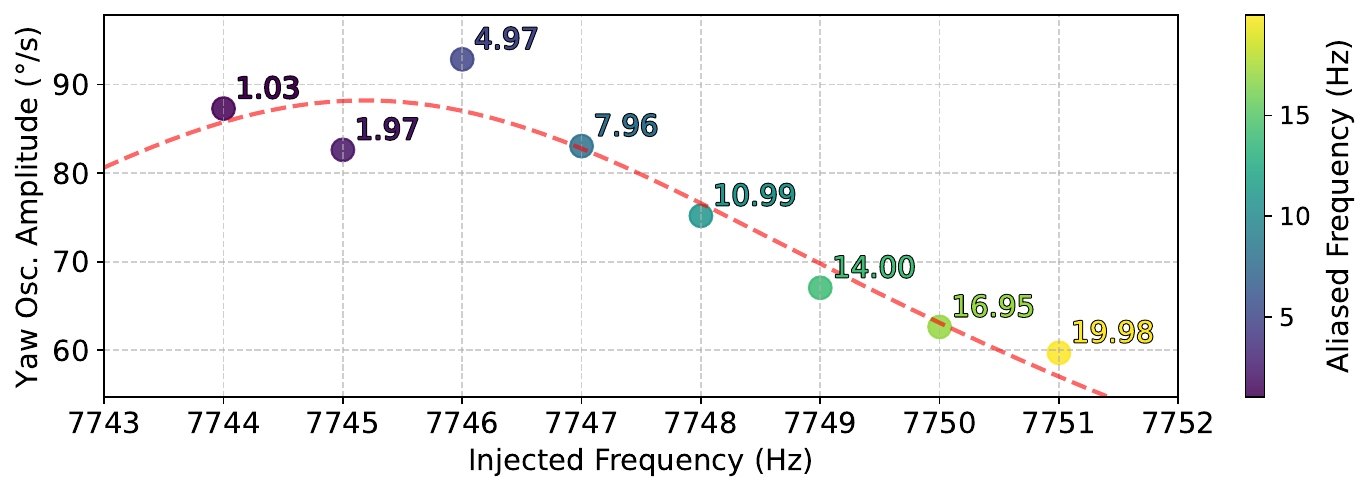}
    \caption{Injected signal frequency vs. yaw oscillation amplitude. Aliased signal frequency shown as changing color. Curve fitted by regression shown as dotted line.}
    \label{fig:freqvamp}
\end{figure}

\section{\rev{Gimbal Oscillation with Directional Biasing}}

\rev{Figure \ref{fig:velocityrepeated} shows 25 consecutive trials of injection at 23232 Hz, 50\% signal power, and clockwise directional biasing in the yaw axis. Pitch, roll, and yaw axis angular velocities are overlaid into a single plot to show we achieve a consistent induced motion with little variation. While some traces show a slight variation in frequency, the amplitude remains consistent, ensuring that the accumulated angular displacement of the gimbal between different trials remains nearly identical as previously shown in Figure \ref{fig:phaseanalysis}.}

\section{Effects of Detuning Injected Frequency}
\label{sec:app_detune}

As the injected frequency $f$ deviates from the natural resonant frequency $f_n$, the aliased frequency increases while the oscillation amplitude decreases. Equation~\eqref{eq:digifreq} shows a linear relationship between $f$ and $f_d$. The amplitude follows from Equation~\eqref{eq:amplitude}, extended with mechanical impedance:
\begin{equation}
\label{eq:impedance}
    Z_m = \sqrt{\left(4 \pi f_n \zeta \right)^2 + \frac{1}{\left( 2 \pi f \right)^2}\left(\left( 2 \pi f_n \right)^2 - \left( 2 \pi f \right)^2 \right)^2},
\end{equation}
where $\zeta$ is the damping ratio. As $|f - f_n|$ increases, $Z_m$ grows, reducing amplitude.

In our evaluated systems, this reduction is moderate, but it becomes important when signal strength is limited or damping is high. This creates a trade-off: larger amplitudes increase attack impact, while higher frequencies yield more stable responses and faster optimization. We model this trade-off by fitting $\zeta$, $F_0$, $m$, and $f_n$ via regression to measured data. The aliased frequency follows a near-linear trend ($R^2=0.99$), enabling accurate prediction. Figure~\ref{fig:freqvamp} confirms that as $f$ moves away from $f_n$, aliased frequency increases while amplitude decreases, and that the fitted model closely matches observations.